# Domain-Adaptive Health Indicator Learning with Degradation-Stage Synchronized Sampling and Cross-Domain Autoencoder

Author

Jungho Choo[1], Hanbyeol Park[1], Gawon Lee[1], Yunkyung Park[2], Hyerim Bae[3*]

[1]Major in Industrial Data Science Engineering, Department of Industrial Engineering, Pusan National University, Busan, 46241, Republic of Korea, {jhchoo, pb104, monago}@pusan.ac.kr

[2]C4ISTAR IPS R&D LIG Nex1, Pangyo, 13488, Republic of Korea, yunkyung.park@lignex1.com

[3*]Department of Data Science, Graduate School of Data Science, Pusan National University, Busan, 46241, Republic of Korea, hrbae@pusan.ac.kr

[3*]Corresponding author

**Abstract.** The construction of high quality health indicators (HIs) is crucial for effective prognostics and health management. Although deep learning has significantly advanced HI modeling, existing approaches often struggle with distribution mismatches resulting from varying operating conditions. Although domain adaptation is typically employed to mitigate these shifts, two critical challenges remain: (1) the misalignment of degradation stages during random mini-batch sampling, resulting in misleading discrepancy losses, and (2) the structural limitations of small-kernel 1D-CNNs in capturing long-range temporal dependencies within complex vibration signals. To address these issues, we propose a domain-adaptive framework comprising degradation stage synchronized batch sampling (DSSBS) and the cross-domain aligned fusion large autoencoder (CAFLAE). DSSBS utilizes kernel change-point detection to segment degradation stages, ensuring that source and target mini-batches are synchronized by their failure phases during alignment. Complementing this, CAFLAE integrates large-kernel temporal feature extraction with cross-attention mechanisms to learn superior domain-invariant representations. The proposed framework was rigorously validated on a Korean defense system dataset and the XJTU-SY bearing dataset, achieving an average performance enhancement of 24.1% over state-of-the-art methods. These results demonstrate that DSSBS improves cross-domain alignment through stage-consistent sampling, whereas CAFLAE offers a high-performance backbone for long-term industrial condition monitoring.

## 1. Introduction

Modern industrial systems are increasingly complex owing to the heightened demands for long-term stability and intensified interdependence among components. Consequently, the failure of a single component can trigger cascading faults, resulting in severe system-level disruptions [1–3]. Prognostic health management (PHM) supports condition-based maintenance strategies by supporting continuous monitoring and health-state assessment [3–5]. A core element of PHM is the health indicator (HI), which quantifies the degree of equipment degradation and aids condition monitoring [2,6–7].

HIs are typically constructed using two main approaches: physical HIs (PHIs) and virtual HIs (VHIs). PHIs are constructed by applying statistical or signal-processing techniques to sensor data [2,8]. Traditional PHIs metrics include root mean square (RMS) [9], kurtosis [10], state-space entropy [11], moving-average cross-correlation [12], and the Gini–Simpson index [13]. These measures offer strong interpretability owing to their clear physical significance [9]. However, their effectiveness often relies on domain expertise and can become unstable across different equipment types or operating conditions, as they typically capture degradation using single or localized statistics [2,6,8]. Kurtosis is highly sensitive to noise and can be distorted by outliers unrelated to degradation, whereas moving-average cross-correlation is vulnerable to external disturbances [9]. These limitations indicate that a single PHI often cannot comprehensively characterize diverse degradation patterns in complex systems [8]. Therefore, VHIs integrate multiple PHIs through dimensionality reduction [8]. Common VHI construction methods, such as kernel principal analysis(KPCA)PCA, support vector machine(SVM), and isometric mapping(ISOMAP )[2], summarize correlated structures and latent degradation patterns into a single index [2,8]. However, KPCA struggles with nonlinear or non-Gaussian data [2,6,8]. Although SVM and ISOMAP can model nonlinear relationships, they require graph construction over all data points, resulting in high computational complexity and sensitivity to hyperparameter selection, increasing overfitting risk [6]. Consequently, conventional VHI methods are often inadequate for stable and generalizable HI construction in complex, real-world environments.

To address the limitations of PHI and VHI, deep learning (DL)-based HI construction has garnered attention. DL-based approaches can capture nonlinear degradation patterns and extract features end-to-end, providing more robust performance across diverse equipment types [6]. These methods are typically categorized as unsupervised and supervised. Unsupervised methods commonly rely on representation distances between healthy and degraded states in run-to-failure (RtF) datasets [13], making HI quality contingent on the ability of the network to extract degradation-related features. Recent architectures have sought to enhance feature extraction by incorporating multi-scale structures or attention mechanisms [8]. However, owing to the absence of sufficient reference signals in unsupervised learning, these methods may misinterpret non-degradation factors—such as sensor noise or spikes as degradation, resulting in latent representations that capture irrelevant characteristics. This misrepresentation can cause the constructed HI to diverge from the actual degradation trend, reducing performance [6,14]. Conversely supervised HI construction methods enhance noise robustness by enforcing monotonic degradation through shape constraint functions (SCFs) within autoencoder frameworks [14–16]. Although supervised models—including CNN- and Transformer-based autoencoders—have demonstrated improved HI performance, they generally assume consistent data distributions between training and testing [8,15,17]. In industrial settings, frequent changes in operating conditions often lead to significant distribution shifts, undermining the effectiveness of monotonic mapping enforced by SCFs and resulting in suboptimal HI construction. Moreover, because SCFs are tailored to specific domain distribution, they cannot be directly transferred across domains, representing a fundamental structural limitation.

To address cross-domain distribution gaps caused by changes in operating conditions, HIs should be constructed with a domain adaptation (DA) framework. DA, a subfield of transfer learning, includes discrepancy-based, adversarial-based, and reconstruction-based approaches, depending on the learning strategy [18]. For example, Chen et al. introduced the transfer quadratic function multi-scale deep convolutional autoencoder (TQFMDCAE), which leverages an inception module and a quadratic SCF to extract domain-specific features [19]. This model mitigates domain shift by aligning distributions using maximum mean discrepancy (MMD). Similarly, Duan et al. incorporated pyramid convolution (PyConv) and Transformer architectures to enhance domain-specific feature representations and improved HI construction by aligning source label across different operating conditions through individual degradation feature alignment [20]. Although these studies advanced HI construction through improved feature extraction, they primarily focus on network architecture and overlook training inefficiencies caused by random mini-batch sampling, where mixed degradation stages can destabilize domain loss and result in misaligned domain signals and oscillatory loss during training. Figure 1 illustrates representative degradation pattern types under different conditions and shows that the degradation signal can be segmented into distinct stages over time; at each stage, the level and variability of observations differ, causing the stage-wise distributions to shift from a healthy state toward a failure state. As shown in Figure 1, degradation data exhibit heterogeneous patterns over time, implying that even for the same equipment, the degradation stage may vary depending on the time point. Therefore, when samples from different degradation stages are randomly mixed within a mini-batch, the domain loss becomes unstable and the alignment signal is distorted, which can exacerbate alignment mismatch. Moreover, most domain adaptation (DA)-based HI models still rely on small-kernel 1D-CNN architectures. However, capturing long-range temporal dependencies is essential for industrial time-series data in which degradation accumulates over long periods, and small-kernel CNNs, due to their limited effective receptive field, cannot sufficiently represent such long-term patterns. As a result, the model produces suboptimal feature representations that fail to fully reflect degradation characteristics. In summary, current DA-based HI construction methods have the following key limitations:

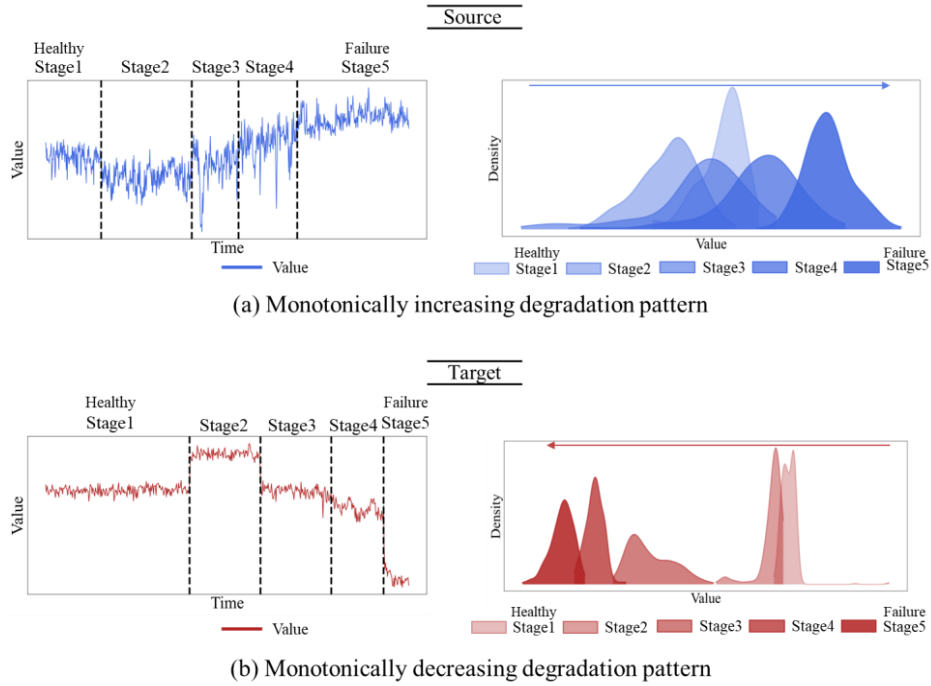

**Figure 1 Diversity in degradation patterns across domains. The equipment in blue displays a monotonically increasing degradation pattern, whereas that in red indicates a monotonically decreasing pattern. In such cases, random batch sampling results in incorrect domain-loss calculations.**

1. Degradation stage mismatched discrepancy loss: Equipment signal distributions differ across degradation stages, each demonstrating distinct statistical patterns [6,8]. Existing DA methods typically use random mini-batch sampling, which mixes source and target samples from various degradation stages within a batch. Calculating discrepancy losses under these mismatched conditions forces the model attempts to align fundamentally different degradation patterns, resulting in misleading alignment signals and ultimately poor HI quality.
2. Limited effective receptive field of 1D-CNN: Although 1D-CNNs are widely used for feature extraction, their small kernels restrict the effective receptive field (ERF), hindering the capture of long-range temporal dependencies in equipment degradation signals [6, 21-22]. To address this, recent studies have introduced multi-scale structures or large-kernel 1D-CNNs—Modern TCN architectures—to expand the ERF and capture broad temporal contexts without excessive network depth [21, 46]. However, these approaches remain underutilized in HI construction. Given that sensor signals often contain complex and long-lasting temporal patterns, extracting long-horizon time-series features is essential for accurately capturing degradation behavior. Therefore, large-kernel based multi-scale feature extraction modules are essential for learning long range degradation temporal patterns across domains [23].

Therefore, this study proposes degradation stage synchronized batch sampling (DSSBS) and the cross-domain aligned fusion large autoencoder (CAFLAE). DSSBS utilizes kernel change point (KCP) detection to identify equipment-specific degradation stages and calculates the MMD loss using samples from the same stage. By synchronizing source and target data sampling within identical degradation stages, DSSBS mitigates spurious domain loss computation resulting from heterogeneous stage mixing. The proposed CAFLAE advances long-horizon temporal feature extraction through a large-kernel parallel modern multi-scale temporal convolution (PMTC) block, surpassing conventional 1D-CNNs by expanding the ERF. Additionally, CAFLAE incorporates a cross-attention module to integrate complementary cross-domain information, enabling higher-quality HI construction. Extensive experiments on a Korean defense system dataset and XJTU-SY bearing dataset demonstrate that the proposed framework delivers superior HI performance under diverse operating conditions. The main contributions of this study are:

1. We propose DSSBS, to the best of our knowledge the first sampling strategy that explicitly addresses domain alignment errors caused by degradation-stage mismatch in randomly constructed mini-batches.

By leveraging KCP-based degradation stage detection, DSSBS ensures that each batch contains source and target samples from the same degradation stage, preventing incorrect domain loss computation.
2. We propose CAFLAE, a DA-based HI construction model that leverages the Parallel Modern multi-scale Temporal Convolution block to strengthen long-horizon time-series feature extraction. CAFLAE enables robust and high-quality HI construction by further adopting a cross-attention module to fuse information across domains.
3. In extensive experiments on the Korean weapon system dataset and the XJTU-SY dataset, the proposed method achieves a 24.1% (+0.138) improvement in comprehensive index (CI) over existing methods.

The remainder of this paper is organized as follows. Section 2 details the proposed DSSBS and CAFLAE. Section 3 outlines the experimental setup and procedures. Section 4 analyzes parameter sensitivity, whereas Section 5 assesses the effectiveness of the proposed methodology through ablation studies. Section 6 concludes and summarizes the findings.

## 2. Methodology

This section presents the overall DSSBS procedure, the CAFLAE architecture, and the end-to-end training process. The proposed workflow of DSSBS and CAFLAE is shown in Figure 2, with notations listed in Table 1.

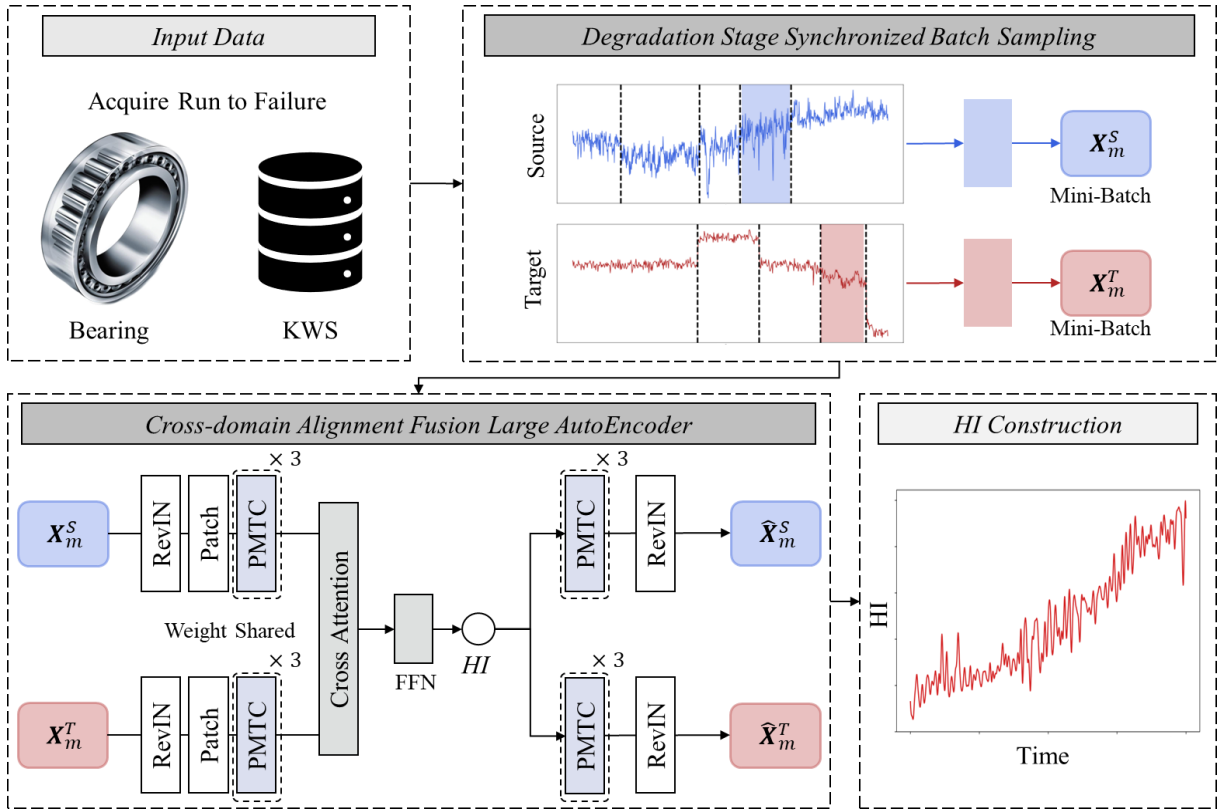

**Figure 2** Overview of the proposed DSSBS and CAFLAE framework. Run-to-failure data are segmented into degradation stages, batch sampling is synchronized by stage, and CAFLAE learns domain-invariant representations to construct HI.

**Table 1. Description of notation**

| Notation | Description |
|---|---|
| $D \in \{S, T\}$ | Set indication Source ($S$) and Target ($T$) domain |
| $X_t^D \in \mathbb{R}^{C \times L \times F}$ | Run to Failure (RtF) time-series tensor of domain $D$ at time $t$, comprising $C$ sensor channels, each with temporal length $L$ and $F$ features. Each channel signal is represented |

| Symbol | Description |
|---|---|
| | as $x_{t,c}^D \in \mathbb{R}^{C \times L \times F}$ and $X_t^D = \{x_{t,c}^D | 1 \le c \le C\}$, where $x_{c,t}^D$ is Channel-wise scalar representation employed for window-wise RMS construction at index $t$ |
| $X_m^D$ | Stage-level segment of domain $D$ obtained by mapping the detected change points in the window index space to the original time axis |
| $\widetilde{X}_t^D$ | RevIN-normalized input of domain $D$ at index $t$ |
| $\widetilde{X}_{emb}^D$ | RevIN-normalized $\widetilde{X}$ with patch embedding |
| $\widehat{X}^D$ | Reconstruction signal in domain $D$ |
| $Y$ | Domain-aware feature obtained by concatenation in domain $D$ |
| $n$ | Sequence length $\{x_t^D\}_{t=1}^N$ in domain $D$ |
| $\omega$ | Window length |
| $N$ | Number of intervals window in domain $D$, typically $N = [n/\omega]$ |
| $\boldsymbol{\mu}_i^D \in \mathbb{R}^C$ | Channel wise RMS feature vector of the $i$-th equal intervals window in domain $D$ |
| $K^D(u,v)$ | RBF kernel similarity between the window feature vectors $\mu_u^D$ and $\mu_v^D$, where $u$ and $v$ represent the indices of two sliding windows in domain $D$. The RBF kernel measures the similarity of two window segments in terms of their feature distributions. |
| $\sigma$ | Bandwidth parameter of the RBF kernel |
| $J^D(a,b)$ | Kernel-based homogeneity cost of segment $[a,b]$ in domain $D$, where $[a,b]$ is start and end index of the segment |
| $M$ | Number of change points in the target domain |
| $c_1, c_2$ | Positive penalty coefficients controlling over-segmentation in the target-stage selection |
| $\widehat{M}$ | Number of change points in the source domain |
| $\tau_0^D, \cdots, \tau_M^D$ | Ordered segmentation boundaries in domain $D$ |
| $\hat{\tau}_M^D$ | Selected optimal change point in domain $D$ |
| $\gamma^D, \beta^D$ | Learnable affine scale and shift parameters of RevIN for domain $D$ |
| $p_i \in \mathbb{R}^P$ | Embedded representation of the $i$-th patch |
| $W_p, b_p$ | Weight matrix and bias for patch projection |
| $\mathbb{K}$ | Kernel size sets for multi-scale depthwise convolutional branches |
| $Q^D, K^D, V^D$ | Query, key, and value projections for domain $D$ |
| $h$ | Number of attention heads |
| $A^{(S \leftarrow T)}, A^{(T \leftarrow S)}$ | Cross-attention outputs |
| $HI_t$ | Scalar health indicator from feed-forward network |
| $SCF(t)$ | Shape constraint function at time $t$ |
| $t_{epoch}$ | Learning epoch index |
| $p$ | Loss term index |
| $\lambda_p^{t_{epoch}}$ | DWA weight for loss term $p$ at epoch $t_{epoch}$ |
| $r_p^{t_{epoch}}$ | Relative loss change rate utilized in dynamic weighted average |
| $Temp$ | Temperature control of the sharpness of dynamic-weighted average weights |

## 2.1 Degradation Stage Synchronized Batch Sampling

Section 2.1 introduces DSSBS, which identifies degradation stages in each domain and synchronizes cross-domain batch sampling within the same degradation stage. The operating mechanism of DSSBS is shown in Figure 3.

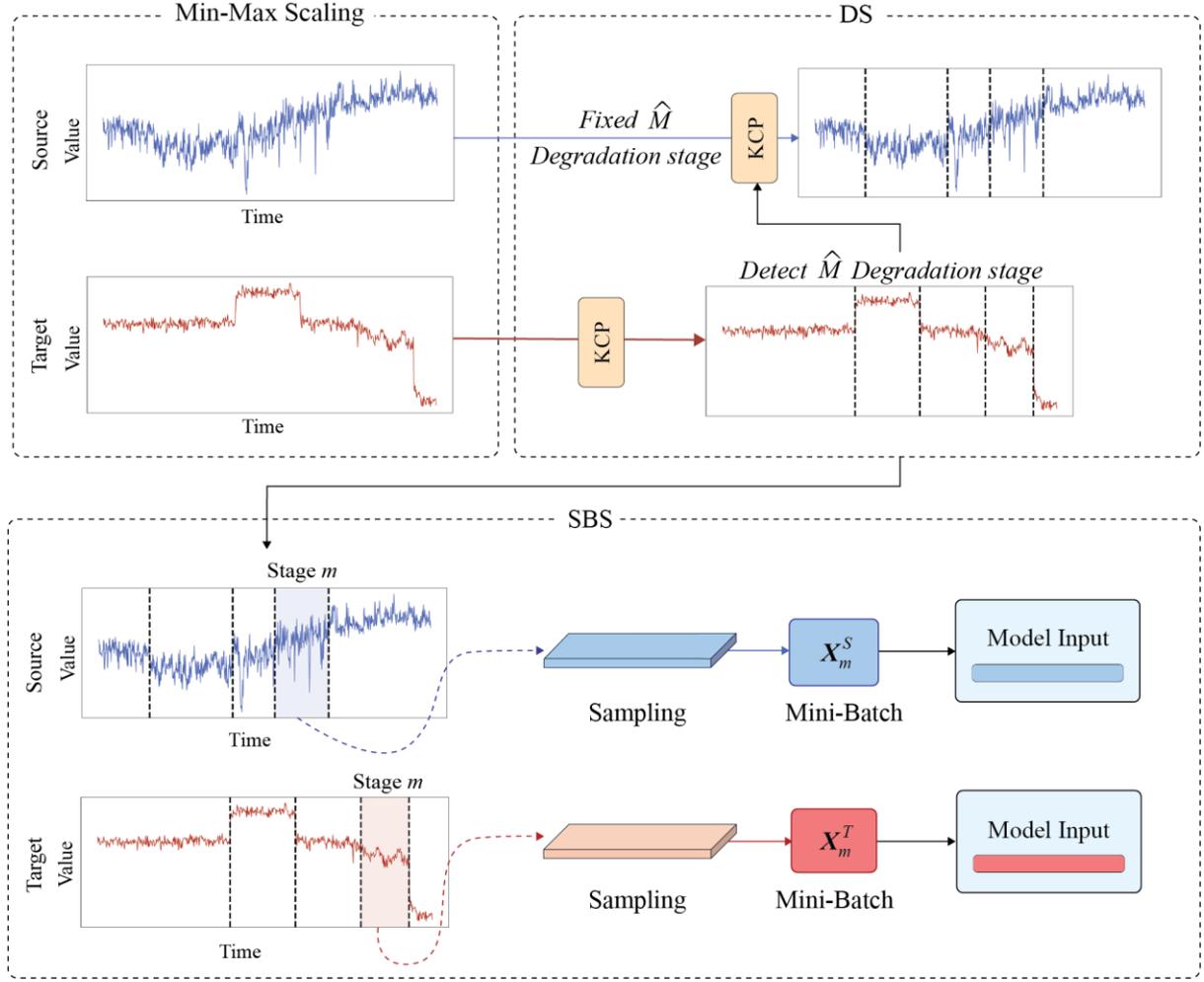

**Figure 3** Overview of the proposed DSSBS process. The RtF signals of the source and target domains are first segmented into degradation stages using kernel change point detection. Each segment is then assigned a stage label, and DSSBS constructs synchronized mini-batches by sampling source and target data from identical degradation stages.

DSSBS partitions RtF sequence into degradation stages using the KCP algorithm, then synchronizes mini-batches such that the source and target samples correspond to the same stage through batch-synchronized sampling. RMS features are extracted from each RtF sensor signal at uniform intervals, and the number of degradation stages, ($M$) is determined based on the target domain. Defining $M$ from the target domain prevents negative transfer resulting from inconsistent stage definitions, which would disrupt stage correspondence and compromise batch-synchronized sampling [24]. For the source domain, DSSBS fixes $\hat{M}$ to the value determined from the target domain and estimates only the change points, preventing batch synchronization from overcompensating for segmentation mismatch, while enabling the model to reduce stage-conditioned distribution shift under the matched degradation-stage condition. The overall procedure is as follows:

1) Min-Max Scaling: The source and target domains are defined as $X^S \in \mathbb{R}^{C \times L \times F}$ and $X^T \in \mathbb{R}^{C \times L \times F}$, respectively. After applying Min–Max scaling to each channel, channel-wise RMS features are computed at a uniform window interval of length $\omega$:

$$\mu_{i,c}^D = \sqrt{\frac{1}{\omega} \sum_{t=(i-1)\omega+1}^{i\omega} \left(x_{c,t}^D\right)^2}, \qquad i = 1, \cdots, N^D, \quad c = 1, \cdots, C, \quad D \in \{S, T\} \qquad (1)$$

where $\mu_{(i,c)}^D$ denotes the RMS value of the $c$-th channel at the $i$-th window in domain $D$ and $N^D = \lfloor L^D/\omega \rfloor$ represents the number of windows obtained by uniformly segmenting the entire time series with window size $\omega$. The RMS feature vector for the $i$-th window is defined as $\boldsymbol{\mu}_i^D = [\mu_{i,1}^D, \cdots, \mu_{i,C}^D]^T \in \mathbb{R}^C$. RMS is used because it amplifies changes in vibration magnitude and is highly sensitive to gradual degradation trends [16]. Unlike the mean, RMS captures both overall energy and degradation severity, aiding in distinguishing degradation stages [6,16,25].

2) Kernel Projection: The uniformly extracted RMS feature vectors are then mapped using a radial basis function (RBF) kernel:

$$K^D(u,v) = \exp\left(-\frac{\|\boldsymbol{\mu}_u^D - \boldsymbol{\mu}_v^D\|_2^2}{2\sigma^2}\right), \quad u,v \in \{1,\cdots,N^D\} \tag{2}$$

The kernel is constructed independently for the source and target domains.

3) Target KCP: KCP is first applied to the target domain to identify degradation stages. Given the uniformly segmented RMS feature sequence $\boldsymbol{\mu}_1^T, \boldsymbol{\mu}_2^T, \cdots, \boldsymbol{\mu}_{N^T}^T$, the kernel-based cost for an interval $[a,b]$ is defined as:

$$J^T(a,b) = \frac{1}{b-a+1}\sum_{u=a}^{b} K^T(u,u) - \frac{1}{(b-a+1)^2}\sum_{u=a}^{b}\sum_{v=a}^{b} K^T(u,v) \tag{3}$$

As described by Garreau and Arlot [27], this cost function quantifies the within-segment homogeneity.

The segmentation on the target domain is represented as:

$$0 = \tau_0^T < \tau_1^T < \cdots < \tau_{M-1}^T < \tau_M^T = N^T \tag{4}$$

The optimal change points in the target domain are identified by minimizing the following objective. Because the number of stages $M$ is unknown, the target domain determines both $M$ and the change points by minimizing a penalized objective:

$$(\widehat{M}, \hat{\tau}^T) = \arg\min_{M,\tau_1^T,\cdots,\tau_{M-1}^T}\left[\sum_{m=1}^{M} J^T(\tau_{m-1}^T+1, \tau_m^T) + \frac{1}{N^T}\left(c_1 \log\binom{N^T-1}{M-1} + c_2 M\right)\right] \tag{5}$$

Where $c_1, c_2 > 0$ represent penalty coefficients to prevent over-segmentation. Finally, the degradation stages of the target domain are represented as follows:

$$\boldsymbol{X}_m^T = \{x_t | t = \hat{\tau}_{m-1}^T \omega + 1, \cdots, \hat{\tau}_m^T \omega\}, \quad m = 1, \cdots, \widehat{M} \tag{6}$$

4) Source KCP with stage-count guarantee: In the source domain, the number of stages $M$ is fixed to match the degradation stages. For $\{\mu_i^S\}_{i=1}^{N^S}$, change points are detected using the cost function $J^S(a,b)$ as follows:

$$J^S(a,b) = \frac{1}{b-a+1}\sum_{u=a}^{b} K^S(u,u) - \frac{1}{(b-a+1)^2}\sum_{u=a}^{b}\sum_{v=a}^{b} K^S(u,v) \tag{7}$$

$$0 = \tau_0^S < \tau_1^S < \cdots < \tau_{\widehat{M}-1}^S < \tau_{\widehat{M}}^S = N^S \tag{8}$$

$$\{\hat{\tau}_1^S, \cdots \hat{\tau}_{\widehat{M}-1}^S\} = \arg\min_{\{\tau_1^S,\cdots,\tau_{\widehat{M}-1}^S\}} \sum_{m=1}^{\widehat{M}} J^S(\tau_{m-1}^S+1, \tau_m^S) \tag{9}$$

By applying $\widehat{M}$ from the target domain to the source domain, we obtain stage partitions in both domains with an identical number of degradation stages:

$$\boldsymbol{X}_m^S = \{x_t | t = \hat{\tau}_{m-1}^S \omega + 1, \cdots, \hat{\tau}_m^S \omega\}, \quad m = 1, \cdots, \widehat{M} \tag{10}$$

The resulting stage intervals are sampled in stage-wise to compute domain loss during training. For equipment with short operating periods, limited samples in certain stages may cause data imbalance across domains. To address this, DSSBS employs a repeated sampling strategy: when a degradation stage lacks sufficient samples in a domain, samples within that stage are repeatedly drawn to balance stage-wise samples.

## 2.2 Cross-domain Alignment Fusion Large AutoEncoder

As shown in Figure 4, CAFLAE learns shared degradation features across source and target domains. The proposed framework adopts a weight-shared autoencoder structure that constructs HIs and reconstructs raw signals for each domain. CAFLAE integrates reversible instance normalization (RevIN), patch embedding, the PMTC block, and cross-attention to fuse domain-specific and domain-invariant representations, constructing a one-dimensional HI from the latent vector. The decoder then reconstructs time-series signals for both domains based on the fused representation. The encoder and decoder architectures are outlined in Sections 2.2.1 and 2.2.2, respectively.

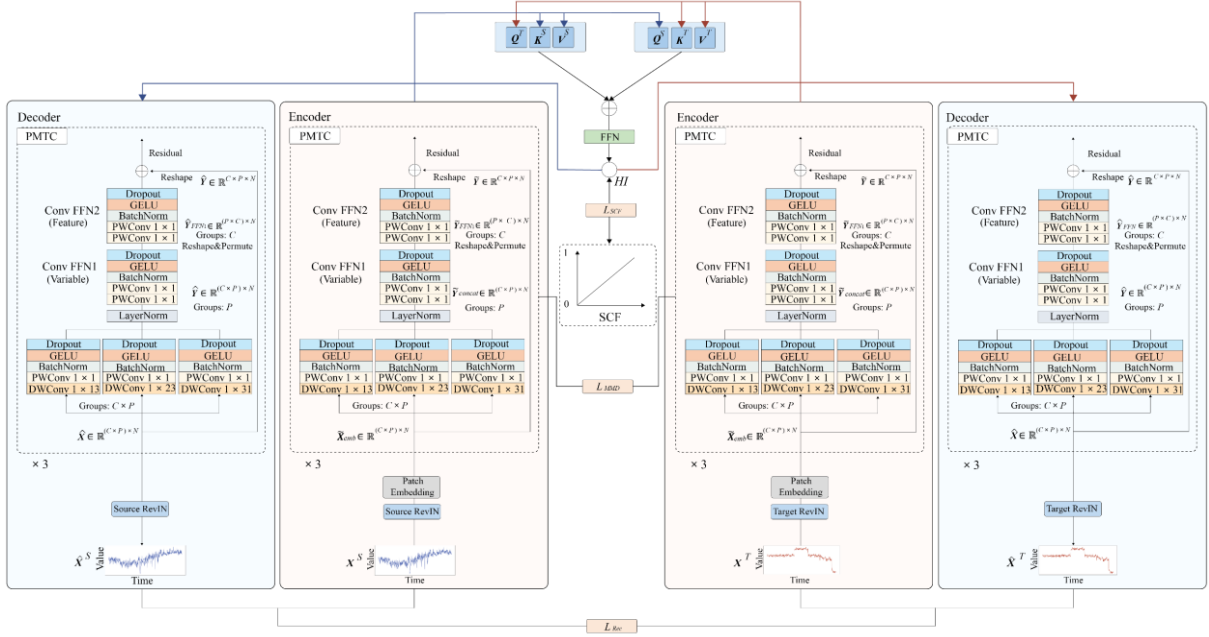

**Figure 4** Overview of the CAFLAE architecture. The model comprises a shared encoder that integrates RevIN, patch embedding, PMTC modules, and cross attention to extract domain-specific and domain invariant features. The decoder reconstructs source and target signals, whereas the HI is supervised through an SCF-based monotonic degradation constraint.

### 2.2.1 CAFLAE Encoder

The CAFLAE encoder extends the architecture of ModernTCN [28] to learn degradation-related features that are transferable across domains. To mitigate the distribution gap between the source and target domains, the encoder employs a weight-shared strategy, using identical parameters for both domains, and is paired with a decoder to reconstruct the input signals.

RevIN normalizes each input time series using its mean and variance, restoring the original distribution with stored statistics after neural processing [29]. To better capture domain-specific statistical characteristics, separate RevIN modules are applied to the source and target domains.

Given an input tensor $X \in \mathbb{R}^{C \times L \times F}$, representing the channel, length, and feature dimensions, the RevIN transform for each domain is defined as:

$$\widetilde{X}_t^D = \gamma^D \odot \frac{X_t^D - \text{Mean}^D}{\sqrt{(\text{Var}^D)^2 + \epsilon}} + \beta^D, \qquad D \in \{S, T\} \tag{11}$$

where $\text{Mean}^D, \text{Var}^D$ denote the mean and variance statistics of the source and target domains, respectively, and $\gamma$ and $\beta$ represent the learnable scaling and shifting parameters, respectively.

To effectively capture localized temporal patterns, patch embedding segments a long time series into shorter patches of length $P$. Given the RevIN-normalized input $\widetilde{X}^D \in \mathbb{R}^{C \times L \times F}$, the $i$-th patch is extracted by a patch operator $\mathcal{P}_i(\cdot)$ as:

$$\widetilde{X}_{emb}^D = \text{patch}(\widetilde{X}_t^D; P, \mathcal{S}) \in \mathbb{R}^{C \times P \times F} \tag{12}$$

Where patch(·) denotes a patch embedding operator, $P$ is a 1D convolution stem with kernel size and stride $S$.

The resulting feature map is then fed into the PMTC encoder block (Figure 4), which models multi-scale temporal dependencies characteristic of degradation processes. Consequently, we adopt depthwise separable convolution (DWConv) [26]. The input vector $\tilde{x}_{emb}$ is reshaped and permuted to $\tilde{x}_{emb} \in \mathbb{R}^{(C \times P) \times L}$, and DWConv is applied with grouped convolutions over $C \times P$ channels using small, medium, and large kernel sizes branch set $\mathbb{K} \in \{13, 23, 31\}$ to capture temporal patterns across varying receptive-field lengths. A subsequent pointwise convolution (PWConv) with a 1×1 kernel fuses kernel information. In DWConv, grouped convolutions over $C \times P$ ensure each channel and feature is processed independently. This DWConv-PWConv combination reduces computational cost while effectively extracting both short- and long-range temporal dependencies. The PMTC operation for each kernel size is expressed as follows:

$$Y_{(\mathbb{K})} = Dropout\left(GELU\left(BatchNorm\left(PWConv\left(DWConv_{(\mathbb{K})}(\tilde{x}_{emb})\right)\right)\right)\right), \quad \tilde{x}_{emb} \in \mathbb{R}^{(C \times P) \times L} \quad (13)$$

The three multi-scale features are concatenated along the channel dimension and normalized to stabilize the representation:

$$Y_{concat} = \text{Concat}(Y_{(13)}, Y_{(23)}, Y_{(31)}), \quad \tilde{Y} = LayerNorm(Y_{concat}), \quad \tilde{Y} \in \mathbb{R}^{(C \times P) \times L} \quad (14)$$

To jointly enhance temporal interactions and inter-channel dependencies, the concatenated feature map $\tilde{Y}$ is further processed using a convolutional feed-forward network (ConvFFN), which adopts the inverted bottleneck design in ModernTCN [28]: a second PWConv expands the hidden channel dimension by a ratio $r$ before projecting it back to the original dimension.

The initial module, ConvFFN1, applies a grouped $1 \times 1$ PWConv with $C$ groups for variable-wise independent feature mixing, followed by another $1 \times 1$ PWConv that expands the hidden channel dimension by a factor of $r$. In this step, the $C \times P$ input channels are divided into $C$ groups, restricting interactions to the $D$ features within each group. A GELU activation [42] and dropout are then applied.

The second module, ConvFFN2, permutes the input into an $P \times C$ configuration, enabling variable interactions from the feature perspective. Setting the number of groups to $P$ partitions the $P \times C$ input channels into $D$ groups, allowing interactions only among the $C$ variables within each group. ConvFFN2 utilizes the same computation as ConvFFN1. Finally, the output of ConvFFN2 is added to the input $\tilde{x}$ through a residual connection, resulting in $Y$:

$$\tilde{Y}_{FFN1} = Dropout\left(GELU\left(PWConv\left(PWConv(\tilde{Y}_{Concat})\right)\right)\right), \quad \tilde{Y}_{Concat} \in \mathbb{R}^{(C \times P) \times L} \quad (15)$$

$$\tilde{Y}_{FFN2} = Dropout\left(GELU\left(PWConv\left(PWConv(\tilde{Y}_{FFN1})\right)\right)\right), \quad \tilde{Y}_{FFN1} \in \mathbb{R}^{(P \times C) \times L} \quad (16)$$

$$Y = (\tilde{Y}_{FFN2} + \tilde{x}), \quad Y \in \mathbb{R}^{C \times P \times L} \quad (17)$$

Let the source and target encoder outputs be $Y^S, Y^T \in \mathbb{R}^{C \times P \times L}$. For each domain, the query, key, and value are computed through domain-specific linear projections not shared across domains. This enables the model to learn domain-specific transformations before cross-domain alignment:

$$Q^D = W_Q^D Y^D, \quad K^D = W_K^D Y^D, \quad V^D = W_V^D Y^D, \quad D \in \{S, T\} \quad (18)$$

where $Q^D, K^D, V^D \in \mathbb{R}^{h \times P \times L}$ denote the projected feature maps for multi-head attention and $h$ represents the number of attention heads.

The query, key, and value of each domain are then input into the cross-attention mechanism. In the source-conditioned direction, the source domain serves as the query, whereas the target domain provides the key and value; the target-conditioned direction reverses these roles:

$$\begin{aligned} A^{(S \leftarrow T)} &= \text{MHSA}(Q^S, K^T, V^T) \\ A^{(T \leftarrow S)} &= \text{MHSA}(Q^T, K^S, V^S) \end{aligned} \tag{19}$$

where $A^{(S \leftarrow T)} \in \mathbb{R}^{C \times P \times L}$ and $A^{(T \leftarrow S)} \in \mathbb{R}^{C \times P \times L}$ represent the cross-attention feature maps for each domain.

To obtain aligned representations, we apply a residual connection and LayerNorm:

$$\begin{aligned} \widetilde{Y}^S &= LayerNorm(Y^S + A^{(S \leftarrow T)}) \\ \widetilde{Y}^T &= LayerNorm(Y^T + A^{(T \leftarrow S)}) \end{aligned} \tag{20}$$

where $\widehat{Y}^S, \widehat{Y}^T \in \mathbb{R}^{C \times P \times L}$ represents the feature maps after the residual connection and LayerNorm.

Next, $\widetilde{Y}^S$ and $\widetilde{Y}^T$ are concatenated along the channel dimension to obtain a domain-aware feature $M$. Finally, $M$ is passed through global average pooling (GAP) in the latent space and through a feed-forward network to construct the HI, $HI_t$, as in Eq. (23).

$$\begin{aligned} Y &= \text{Concat}(\widetilde{Y}^S, \widetilde{Y}^T), \qquad Y \in \mathbb{R}^{2C \times P \times L} \\ HI_t &= FFN(GAP(Y)) \end{aligned} \tag{21}$$

where $HI_t$ represents a one-dimensional latent representation.

### 2.2.2 CAFLAE Decoder

The CAFLAE decoder reconstructs the raw signals for each domain from the HI extracted by the encoder. The HI is first projected through a PWConv to match the channel dimension of the decoder input:

$$H^D = \text{PWConv}(HI_t), \qquad D \in \{S, T\} \tag{22}$$

where $H^D$ denotes the channel-aligned latent representation obtained by projecting $HI_t$ though a domain-specific PWConv. Finally, inverse RevIN is applied to restore the original distribution of the input time series, as shown in Eq. (25):

$$\widehat{Y}^D = \text{PMTC}(H^D), \qquad D \in \{S, T\} \tag{23}$$

where $\widehat{Y}^D$ represents the reconstructed multi-scale temporal feature map produced by the PMTC block with large-kernel temporal filters. The variable-axis and time-axis ConvFFN modules refine the reconstructed feature map. Residual connections and LayerNorm stabilize the reconstructed signal:

$$\widehat{Y}^D_{\text{FFN}} = LayerNorm\left(\widehat{Y}^D + ConvFFN(\widehat{Y}^D)\right), \qquad D \in \{S, T\} \tag{24}$$

where $\widehat{Y}^D_{\text{ffn}}$ represents the temporal representation obtained after applying the variable-axis and time-axis ConvFFN modules followed by a residual connection and LayerNorm. To restore the original temporal resolution, the decoder stacks three multi-scale transposed-convolution paths with kernel sizes $\mathbb{K}$. Each path progressively upsamples the sequence, while capturing multi-scale temporal variations:

$$\widehat{Y}^D_{(\mathbb{K})} = Dropout(GELU(\text{BatchNorm}(\text{PWConv}(\text{TransDWConv}_{(\mathbb{K})}(\widehat{Y}^D_{\text{FFN}}))))) \tag{25}$$

where $\widehat{Y}^D_{(\mathbb{K})}$ denotes the upsampled features obtained from PMTC, mirroring the multi-scale branches of the encoder. Aggregating the full transposed-convolution pipeline, the decoded signal before inverse normalization is expressed as:

$$\widehat{X}^D_{Dec} = \text{Dec}(\widehat{Y}^D) \tag{26}$$

where $\text{Dec}(\cdot)$ denotes the entire decoder stack comprising the multi-scale transposed convolution paths and reconstruction pointwise projections. Finally, domain-specific inverse RevIN restores the original statistics of each domain:

$$\widehat{X}^D = (\frac{\widetilde{X}^D_{Dec} - \beta^D}{\gamma^D}) \odot \sqrt{(\text{Var}^D)^2 + \epsilon} + \text{Mean}^D \qquad (27)$$

where $\text{Mean}^D$ and $\text{Var}^D$ represent the domain-specific RevIN statistics stored during encoding and are utilized to recover the original distribution of each domain, ensuring that the reconstructed source and target sequences remain statistically consistent with their corresponding raw signals.

## 2.3 Training Procedure

CAFLAE is trained with an SCF as defined by Chen et al. [15]:

$$SCF(t) = \left(\frac{1}{t_n^\alpha} \times t^\alpha\right) + 1 \qquad (28)$$

where $t_n$ denotes the life cycle index of an RtF sequence, $t$ denotes the current time index, and $\alpha$ controls the curvature of the SCF, which determines how the rate of increase in SCF, thereby adjusting the growth profile of the constraint.

To mitigate the domain shift between the source and target domains, we employ the MMD loss:

$$L_{MMD} = \| \frac{1}{N^S} \sum_{i=1}^{N^S} \phi\left(GAP(f_i^S)\right) - \frac{1}{N^T} \sum_{j=1}^{N^T} \phi(GAP(f_j^T)) \|_\mathcal{K}^2 \qquad (29)$$

where $f^S$ and $f^T$ denote the outputs of the PMTC block, and $\mathcal{K}$ denotes the reproducing kernel Hilbert space.

The SCF loss encourages $HI_t$ to follow the predefined SCF:

$$L_{SCF} = \frac{1}{N} \sum_{t=1}^{N} \| HI_t - SCF_t \|_2^2 \qquad (30)$$

The reconstruction loss is defined as follows:

$$L_{REC} = \frac{1}{N} \sum_{t=1}^{N} \| X_t^S - \widehat{X}_t^S \|_2^2 + \| X_t^T - \widehat{X}_t^T \|_2^2 \qquad (31)$$

Overall, the SCF loss $L_{SCF}$ enforces a monotonic degradation pattern. The domain loss $L_{MMD}$, computed from the PMTC outputs, reduces the discrepancy between source and target feature distributions. The reconstruction loss $L_{REC}$ minimizes the difference between reconstructed and original raw signals in both domains.

For joint optimization, we adopt dynamic weight averaging (DWA) as in Liu et al. [31]. DWA dynamically adjusts the relative importance of each loss term by tracking changes in loss magnitude over time, eliminating manual tuning. Specifically, task weights are updated based on the ratio of losses across consecutive epochs. Let $L_p^{t_{epoch}}$ denote the loss for objective $p \in \{1,2,3\}$ at epoch $t$. As described in Liu et al. [31], the weight update is determined by the change rate $r_p^{t_{epoch}}$, defined by the loss ratio between successive epochs (Eq. (34)).

$$r_p^{t_{epoch}} = \frac{L_p^{t_{epoch}-1}}{L_p^{t_{epoch}-2}}, \qquad p \in \{1,2,3\} \qquad (32)$$

Based on the change rate in Eq. (34), a softmax function computes the weight of each loss term [31]:

$$\lambda_p^{t_{epoch}} = \exp\left(\frac{r_k^{t_{epoch}}/Temp}{\sum_p \exp\left(r_a^{t_{epoch}}/Temp\right)}\right), \qquad p \in \{1,2,3\} \qquad (33)$$

where $Temp$ controls the sharpness of the weight distribution, $p$ indexes the three loss terms, and $r_p$ denotes the loss change rate. Finally, the total loss is constructed as follows [31]:

$$L_{TOT}^{te} = \lambda_1^{te} L_{SCF} + \lambda_2^{te} L_{MMD} + \lambda_3^{te} L_{REC} \tag{34}$$

## 3. Experiments

This section outlines the experimental protocol and results of the proposed methodology. Section 3.1.1 introduces the Korea Weapon System (KWS) dataset, and Section 3.1.2 describes the XJTU-SY bearing dataset. Section 3.2 summarizes the HI evaluation metrics and provides additional analyses, including loss-curve inspection for training stability, t-SNE visualization to assess domain alignment before and after adaptation, and an effective receptive field (ERF) analysis to examine the impact of large kernels in CAFLAE. Section 3.3 details the experimental settings, and Section 3.4 reports the experimental results.

### 3.1 Data Description
#### 3.1.1 KWS Dataset

The KWS dataset comprises data from the armed forces of the Republic of Korea. The cooling system, essential for maintaining 24-hour operational readiness, is divided into two conditions: Refrigerant Compressor 1 (RC1) and Air Compressor 2 (AC2), both sampled at a 360-s operational cycle. The training and testing splits are listed in Table 2.

#### 3.1.2 XJTU-SY Dataset

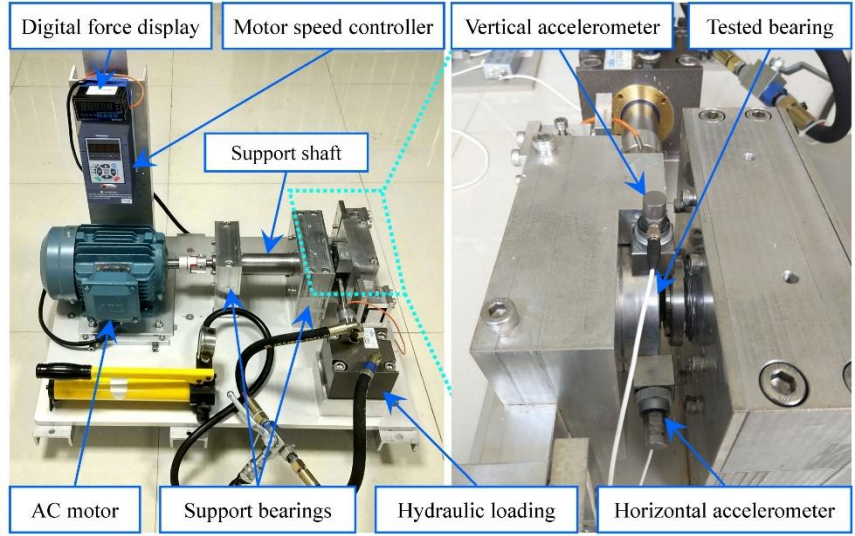

Figure 5 Experimental platform of the XJTU-SY accelerated bearing life test

The XJTU-SY dataset is an accelerated bearing life test platform developed by Xi'an Jiaotong University. The test rig comprises an AC motor, a speed controller, a rotating shaft, a hydraulic loading system, and a bearing test set. LDK UER204 rolling bearings were used, with two PCB 352C33 accelerometers mounted horizontally and vertically on the bearing housing to capture vibration signals [32].

The dataset was collected under three operating conditions:

- Condition 1: 2100 rpm, 12 kN load
- Condition 2: 2250 rpm, 11 kN load
- Condition 3: 2400 rpm, 10 kN load

Vibration signals were sampled every minute at 25.6 kHz for 1.28 s. To satisfy bearing characteristic frequency analysis and sampling density requirements, each sample was normalized to 4096 points [16].

## 3.2 Train and Test Set Spilt

The train/test splits for the same-condition and cross-condition settings are listed in Tables 2 and 3. In the same-condition experiments, both training and testing occurred within the same operating condition. For cross-condition experiments, the source domain included SCF, whereas the target domain was unlabeled; therefore, the HI for the target test set was generated using SCF learned from the source domain. For the KWS dataset, splits were based on the temporal order of the RtF sequences, with earlier sequences for training and later sequences for testing. For the XJTU-SY dataset, the same-condition train/test split followed the protocol in Duan et al. and Li et al. [14,16]. For cross-condition experiments, this protocol was extended to define cross-domain HI construction tasks by assigning different operating conditions to different domains [14,16,22,33]. Additional cross-condition configurations, where the source domain contained only a limited number of labeled bearings, were grouped as Cross Condition-b; their experimental setups and results are provided in the Appendix [7,16,22,33–37].

**Table 2** Dataset configurations for KWS under the same and cross conditions

| Type | Test set | Train set | | Condition | Models |
|---|---|---|---|---|---|
| | | Source (W/ Label) | Target (W/O Label) | | |
| Same Condition | RC1-2, RC1-3 | RC1-1 | | 1→1 | DCAE |
| | AC2-2 | AC2-1 | | 2→2 | SMASE TDCAE |
| Cross Condition | RC1-2, RC1-3 | AC2-1 | RC1-1 | 2→1 | TQFMDCAE |
| | AC2-2 | RC1-1 | AC2-1 | 1→2 | HCPTSCAE CAFLAE |

**Table 3** Dataset configurations for XJTU-SY under same and cross conditions

| Type | Test set | Train | | Condition | Models |
|---|---|---|---|---|---|
| | | Source (W/ Label) | Target (W/O Label) | | |
| Same Condition | B1-1, B1-2 | B1-3, B1-5 | | 1 → 1 | DCAE |
| | B2-3, B2-4, B2-5 | B2-1, B2-2 | | 2 → 2 | SMSAE |
| | B3-5 | B3-3 | | 3 → 3 | TDCAE |
| Cross Condition | B1-1, B1-2 | B2-1, B2-2 | B1-3, B1-5 | 2 → 1 | TQFMDCAE |
| | B2-3, B2-4, B2-5 | B1-3, B1-5 | B2-1, B2-2 | 1 → 2 | HCPTSCAE |
| | B3-5 | B1-1, B1-2 | B3-3 | 1 → 3 | CAFLAE |

## 3.3 Evaluation Metric

To evaluate HI construction performance, we employed widely recognized HI metrics, including monotonicity (Mon), correlation (Cor), robustness (Rob), and the comprehensive index (CI) [6,15,17].

$$Mon(HI) = \frac{1}{3}\left(\frac{|\sum_t S(dHI_{MA}(t,1))|}{N-1} + \frac{|\sum_t S(dHI_{MA}(t,3))|}{N-2} + \frac{|\sum_t S(dHI_{MA}(t,3))|}{N-3}\right) \quad (35)$$

$$Cor(HI, HI_{MA}) = \frac{|N\sum_{t=1}^{N} HI_{MA}(t)t - (\sum_{t=1}^{N} HI_{MA}(t)(\sum_{t=1}^{N} t))|}{\sqrt{[N\sum_{t=1}^{N} HI_{MA}(t)^2 - (\sum_{t=1}^{N} HI_{MA}(t))^2][N\sum_{t=1}^{N} t^2 - (\sum_{t=1}^{N} t)^2]}} \quad (36)$$

$$Rob(HI) = \exp\left(-\frac{\xi}{N}\sum_{t=1}^{N}\left|\frac{HI_R(t)}{\max(HI_{MA}) - \min(HI_{MA})}\right|\right) \quad (37)$$

$$CI = 0.4 * Mon(HI) + 0.3 * Cor(HI, HI_{MA}) + 0.3 * Rob(HI) \quad (38)$$

$$HI_R = HI - HI_{MA} \quad (39)$$

where $HI_{MA}$ denotes the moving-average trend of the constructed HI and $HI_R$ denotes the residual component not explained by the moving-average trend.

Mon evaluates whether the HI consistently increases or decreases over time using a three-point operator. Cor measures the relationship between HI and time using the Pearson correlation coefficient, reflecting the clarity of degradation progression. Rob quantifies smoothness by evaluating the variance of residual fluctuations. The exponential term includes a scaling coefficient set to $\xi = 2$, as in Qin et al. [17]. CI combines Mon, Cor, and Rob with weights of 0.4, 0.3, and 0.3, respectively [6].

### 3.4 Experimental Setting

The experimental setup is outlined in Section 3.4.

**Table 4 Description of HI construction baseline models**

| Same condition | Cross condition | Model | Description |
|---|---|---|---|
| √ | | DCAE [15] | A 1D CNN-based autoencoder architecture composed of multiple convolution and pooling layers in the encoder, as well as transposed convolution and upsampling layers in the decoder. |
| √ | | SMSAE [17] | An autoencoder comprising multi-head self-attention modules followed by a multilayer perceptron, applied to both the encoder and decoder. |
| √ | | TDCAE [16] | An enhanced convolutional autoencoder incorporating Transformer encoding. The encoder comprises 1D convolution layers, batch normalization, and Transformer blocks, whereas the decoder utilizes transposed convolution layers. |
| | √ | TQFMDCAE [19] | A domain adaptation autoencoder based on the inception module and MK-MMD. The encoder comprises inception modules, whereas the decoder is implemented using GAP and transposed convolution layers. |
| | √ | HCPTSCAE [25] | A domain adaptation autoencoder that leverages PyConv and working condition aware feature alignment. The encoder employs PyConv, 2D convolution, batch normalization, and Transformer encoding, whereas the decoder is constructed using transposed convolution layers. |

We evaluated the HI construction performance of the models listed in Table 4, reporting average results over 10 independent runs. For fair comparison, baseline methods are listed in Table 5, including both supervised HI construction models under same-condition settings and cross-condition HI construction models.

**Table 5 Hyperparameter search space for the neural network architectures, including encoder/decoder layer settings, Transformer dimensions, and cross attention configurations**

| Methods | Hyperparameters | |
|---|---|---|
| | Model architecture | Model Training |
| DCAE | $N_e = \{5,6\}; N_d = \{5,6\};$ | $\alpha = \{1,2\}; \lambda_1 = \{0.1, 0.5, 1.0\};$ $\lambda_2 = \{0.1, 0.5, 1.0\}$ |
| SMSAE | $N_e = \{5,6\}; N_d = \{5,6\}; d = \{32, 64\},$ | $\alpha = \{1,2\}; \lambda_1 = \{0.1, 0.5, 1.0\};$ $\lambda_2 = \{0.1, 0.5, 1.0\}$ |
| TDCAE | $N_e = \{5, 6\}; N_d = \{5, 6\}; d = \{64, 128\}$ | $\alpha = \{1,2\}; \lambda_1 = \{0.1, 0.5, 1.0\};$ $\lambda_2 = \{0.1, 0.5, 1.0\}$ |
| TQFMDCAE | $N_e = \{3\}; N_d = \{4\}; k = \{3, 5, 7\}$ | $\alpha = \{1,2\}; \lambda_1 = \{0.1, 0.5, 1.0\};$ $\lambda_2 = \{0.1, 0.5, 1.0\}; \lambda_3 = \{0.1, 0.5, 1.0\}$ |
| HCPTSCAE | $N_e = \{3\}; N_d = \{3\}; d = \{32, 64\};$ | $\alpha = \{1,2\}; \lambda_1 = \{0.1, 0.5, 1.0\};$ |

|  |  |  |  |
|---|---|---|---|
| CAFLAE | $k = \{3, 5, 7\}$ $N_e = \{2,3\}; N_d = \{2,3\}; d = \{32, 64\}, C = \{32, 64\}; k = \{13, 23, 31\}$ $r = \{2, 4\}$ | $\lambda_2 = \{0.1, 0.5, 1.0\}; \lambda_3 = \{0.1, 0.5, 1.0\},$ $\lambda_4 = \{0.1, 0.5, 1.0\}; \lambda_5 = \{0.1, 0.5, 1.0\}$ $\alpha = \{1,2\}; penalty = \{10, 20, 30\};$ $kernel = RBF$ |

The search ranges of neural network architectures and training hyperparameters for each model are listed in Table 5, where $N_e$ and $N_d$ denote the number of convolutional layers in the encoder and decoder, respectively; $d$ denotes the Transformer encoder dimension; and $c$ denotes the dimension of the cross-attention module.

## 3.5 Experimental Results

The effectiveness of the proposed CAFLAE is evaluated from three perspectives. (1) The quality of the constructed HI is evaluated using standard HI evaluation metrics and visualizations to validate whether the HI consistently reflects degradation progression and serves as a reliable indicator for condition monitoring. (2) Training stability is analyzed using training curves and the PI-Control metric, whereas domain alignment performance is evaluated by visualizing source and target features before and after training using t-SNE. (3) The capability of PMTC in CAFLAE to capture long-horizon temporal characteristics is examined through ERF analysis.

### 3.5.1 HI Construction Experimental Results

In this section, we evaluate the constructed HI using the HI performance metrics.

**Table 6 Comparison of HI construction performance on the KWS dataset**

| Test | Method | Metric |  |  |  |
|---|---|---|---|---|---|
|  |  | Mon | Cor | Rob | CI |
| RC1-2 | DCAE | 0.3850 | 0.5958 | 0.7838 | 0.5679 |
|  | SMSAE | 0.3584 | 0.7220 | 0.7617 | 0.5885 |
|  | TDCAE | 0.3834 | 0.5442 | 0.7877 | 0.5530 |
|  | TQFMDCAE | **0.5493** | **0.8931** | 0.6905 | <u>0.6948</u> |
|  | HCPTSCAE | <u>0.5167</u> | 0.7252 | <u>0.7944</u> | 0.6626 |
|  | CAFLAE | 0.5055 | <u>0.8719</u> | **0.9226** | **0.7406** |
| RC1-3 | DCAE | 0.4064 | <u>0.5975</u> | 0.7822 | 0.5765 |
|  | SMSAE | 0.4101 | 0.2704 | 0.7680 | 0.4756 |
|  | TDCAE | 0.4049 | 0.5384 | 0.7796 | 0.5573 |
|  | TQFMDCAE | 0.4858 | 0.4286 | 0.8068 | 0.5605 |
|  | HCPTSCAE | <u>0.4882</u> | 0.5927 | <u>0.8297</u> | <u>0.6220</u> |
|  | CAFLAE | **0.5069** | **0.8291** | **0.9132** | **0.7416** |
| AC2-2 | DCAE | 0.4054 | 0.2499 | 0.7802 | 0.4712 |
|  | SMSAE | 0.4083 | 0.3082 | 0.7735 | 0.4878 |
|  | TDCAE | 0.4028 | 0.1730 | 0.7825 | 0.4478 |
|  | TQFMDCAE | <u>0.5043</u> | 0.1931 | 0.8053 | 0.5012 |
|  | HCPTSCAE | 0.4960 | <u>0.3682</u> | <u>0.8373</u> | <u>0.5600</u> |
|  | CAFLAE | 0.4965 | **0.8655** | **0.9029** | **0.7291** |

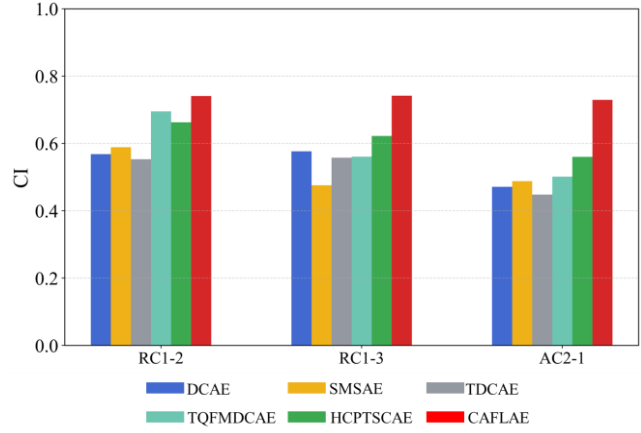

**Figure 6 Comprehensive index of different KWS HI**

The HI construction performance of same-condition and cross-condition models on the KWS dataset is listed in Table 6. For RC1-2, the same-condition models achieved CI scores of DCAE 0.5679, SMSAE 0.5885, and TDCAE 0.5530. Cross-condition models yielded higher CI values, reaching 0.6948 for TQFMDCAE and 0.6626 for HCPTSCAE. CAFLAE achieved the highest CI score of 0.7406. These results indicate that same-condition models demonstrate lower HI performance, whereas cross-condition learning provides clear advantages in robustness and generalization. The performance gap in RC1-2 suggests that incorporating heterogeneous operating conditions results in more reliable degradation modeling. CAFLAE further reinforces this trend by delivering the most stable and accurate HI construction across all benchmarks.

A similar trend was observed in RC1-3, where cross-condition models outperformed same-condition models. The advantage of cross-condition learning became more evident under varying operating conditions. For example, when shifting from RC1-3 to AC2-2, the CI of same-condition models dropped significantly, with SMSAE decreasing from 0.6220 to 0.4756. Conversely, cross-condition models experienced substantially smaller performance declines: CI decreased was approximately 0.0593 for TQFMDCAE and 0.0629 for HCPTSCAE. CAFLAE achieved the best performance across all scenarios. Even in AC2-2, where same-condition models underperformed, CAFLAE achieved a CI of 0.7291, surpassing HCPTSCAE by 0.1691. The proposed framework yielded an average improvement of 24.1% over existing methods, demonstrating its effectiveness in constructing high-quality HI.

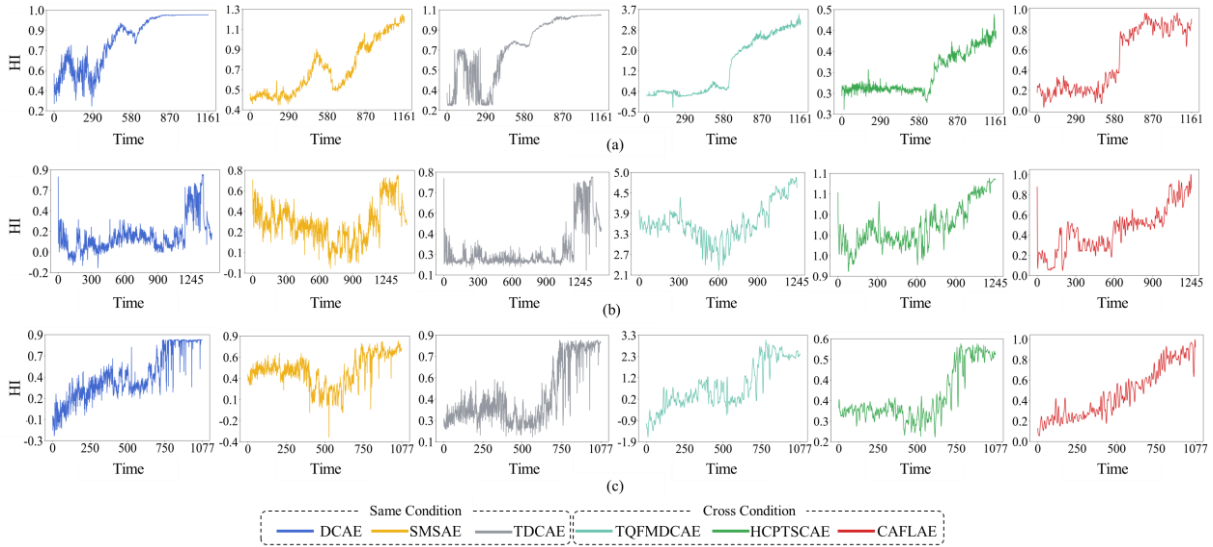

**Figure 7 Visualization of the constructed HIs on the KWS dataset for same condition and cross condition settings across (a) RC1-2, (b) RC1-3 and (c) AC2-2. Each subplot compares the temporal HI trends produced by DCAE, SMSAE, TDCAE, TQFMDCAE, HCPTSCAE, and the CAFLAE**

The HI constructed for each test case, illustrating temporal degradation trends and noise characteristics as the equipment approaches failure, is shown in Figure 7. The x-axis and y-axis denote time and HI values, respectively. Several limitations are apparent in the HI generated by same-condition models. For DCAE, although a degradation trend is visible in RC1-2, the HI reaches the failure threshold prematurely, producing misleading monitoring signals. In RC1-3, the early-stage degradation pattern is unclear, and the HI drops abruptly at failure, resulting in an inaccurate trajectory. In AC2-2, the HI displays a failure-like behavior before the actual event, again generating misleading monitoring information. SMSAE demonstrates a smoother degradation trend than DCAE in RC1-2; however, its HI spans [0.3, 1.3], making it impractical for monitoring [38]. In RC1-3, the HI decreases near failure similarly to DCAE, whereas in AC2-2, it lacks a consistent degradation pattern. For TDCAE, the HI is unstable early in RC1-2 and falsely signals failure later. In RC1-3, the HI decreases rapidly before failure, mirroring SMSAE and indicating unreliable failure characterization. AC2-2 also displays premature failure patterns and significant noise, highlighting limited robustness.

Among cross-condition models, TQFMDCAE is less sensitive to noise and maintains consistent degradation trends across RC1-2, RC1-3, and AC2-2. Furthermore, HCPTSCAE demonstrates relatively stable degradation progression. However, both models generate HI values outside the expected [0, 1] range, which may mislead condition monitoring. For example, TQFMDCAE produces HI values from 0.5 to 3.7 in RC1-2, limiting its practical applicability [38]. Conversely, CAFLAE maintains HI values within [0, 1] across RC1-2, RC1-3, and AC2-2, indicating high suitability for condition monitoring [38]. In RC1-2, the HI increases monotonically, converging to 1 at failure. In RC1-3, the HI increases linearly, reaching 1 precisely at failure without premature alarms, unlike DCAE and TDCAE. AC2-2 demonstrates a similarly stable, increasing trend, demonstrating consistent modeling even under changing conditions. Therefore, the proposed framework can construct stable and high-quality HI even with limited failure history.

**Table 7 Comparison of HI construction performance on the XJTU-SY Dataset**

| Test | Method | Metric | | | |
| --- | --- | --- | --- | --- | --- |
| | | Mon | Cor | Rob | CI |
| B1-1 | DCAE | 0.5494 | 0.8343 | 0.5003 | 0.6202 |
| | SMSAE | 0.5524 | <u>0.8617</u> | 0.5659 | 0.6523 |
| | TDCAE | 0.6808 | 0.8571 | 0.5184 | 0.6880 |
| | TQFMDCAE | <u>0.5643</u> | 0.8182 | <u>0.9331</u> | <u>0.7571</u> |
| | HCPTSCAE | 0.5460 | 0.8593 | 0.8475 | 0.7365 |
| | CAFLAE | **0.5702** | **0.8634** | **0.9451** | **0.7706** |
| B1-2 | DCAE | 0.5390 | 0.8255 | 0.6400 | 0.6553 |
| | SMSAE | 0.5526 | <u>0.8742</u> | 0.6790 | 0.6870 |
| | TDCAE | <u>0.6484</u> | **0.9024** | 0.6683 | 0.7306 |
| | TQFMDCAE | 0.5701 | 0.8520 | <u>0.8739</u> | <u>0.7578</u> |
| | HCPTSCAE | 0.5348 | 0.8289 | 0.8361 | 0.7134 |
| | CAFLAE | **0.6501** | 0.8401 | **0.8913** | **0.7794** |
| B2-3 | DCAE | 0.5096 | 0.8263 | 0.4167 | 0.5767 |
| | SMSAE | **0.5210** | **0.8600** | 0.4679 | 0.6068 |
| | TDCAE | <u>0.5165</u> | <u>0.8480</u> | 0.4432 | 0.5939 |
| | TQFMDCAE | 0.4886 | 0.8422 | <u>0.8630</u> | <u>0.7076</u> |
| | HCPTSCAE | 0.4776 | 0.8356 | 0.7270 | 0.6598 |
| | CAFLAE | 0.5046 | 0.8257 | **0.9225** | **0.7263** |
| B2-4 | DCAE | 0.6204 | 0.7738 | 0.3276 | 0.5786 |
| | SMSAE | 0.6041 | 0.8029 | 0.4208 | 0.6087 |
| | TDCAE | 0.6561 | 0.7554 | 0.3692 | 0.5998 |
| | TQFMDCAE | <u>0.6753</u> | 0.8449 | **0.9305** | <u>0.8027</u> |
| | HCPTSCAE | 0.6152 | <u>0.8500</u> | <u>0.8767</u> | 0.7641 |
| | CAFLAE | **0.7729** | **0.8880** | 0.8066 | **0.8176** |
| B2-5 | DCAE | 0.5245 | 0.8231 | 0.5492 | 0.6215 |
| | SMSAE | **0.5435** | **0.9090** | 0.6869 | 0.6962 |
| | TDCAE | 0.5269 | 0.8364 | 0.6585 | 0.6592 |
|   | TQFMDCAE | 0.5234 | <u>0.8621</u> | 0.8951 | **0.7365** |
|   | HCPTSCAE | 0.4900 | 0.8181 | <u>0.9076</u> | 0.7137 |
|   | CAFLAE | <u>0.5429</u> | 0.7666 | **0.9469** | <u>0.7312</u> |
| B3-5 | DCAE | 0.5143 | 0.4235 | 0.5253 | 0.4903 |
|   | SMSAE | 0.5679 | **0.7882** | 0.7714 | <u>0.6950</u> |
|   | TDCAE | **0.6301** | <u>0.6935</u> | 0.7717 | 0.6916 |
|   | TQFMDCAE | 0.5548 | 0.6781 | 0.8886 | 0.6919 |
|   | HCPTSCAE | 0.5191 | 0.4704 | <u>0.9082</u> | 0.6212 |
|   | CAFLAE | <u>0.6036</u> | 0.6764 | **0.9625** | **0.7331** |

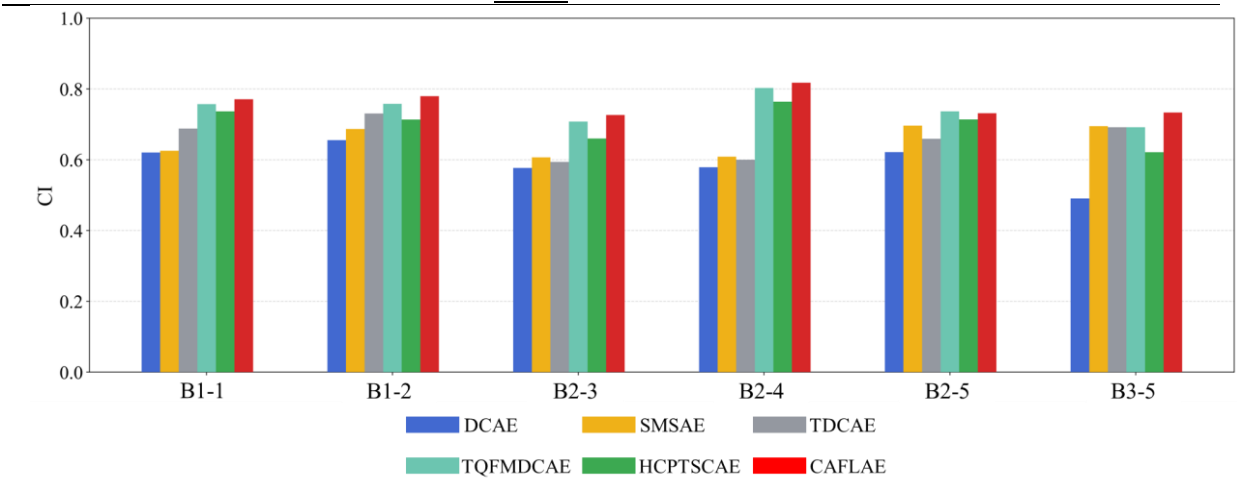

**Figure 8 Comprehensive index of different XJTU-SY HIs**

The HI construction performances on the XJTU-SY dataset under same- and cross-condition settings are listed in Table 7. For B1-1, DCAE achieved the lowest CI (0.6202) in the same-condition setting, whereas TDCAE reached the highest (0.6889). Cross-condition models generally achieved higher scores, with HCPTSCAE and TQFMDCAE reaching CIs 0.7134 and 0.7571. CAFLAE achieved the highest CI (0.7706). These results demonstrate that single-condition training limits generalization, whereas cross-condition learning yields more robust degradation representations. A similar trend emerged in B1-2: DCAE remained less effective under the same-condition setting, whereas cross-condition methods maintained higher CIs. CAFLAE achieved a CI of 0.7794, outperforming TQFMDCAE by 0.0216, reflecting enhanced stability and accuracy under distribution shifts.

This trend intensified in B2-3 and B2-4, where same-condition model CIs dropped significantly, highlighting the limitations of single-condition training. Conversely, TQFMDCAE achieved CIs of 0.7070 and 0.8027, respectively, and HCPTSCAE reached 0.6598 and 0.7641, demonstrating strong cross-condition performance. CAFLAE maintained the highest performance with CIs of 0.7263 and 0.8176, demonstrating superior robustness under condition changes. In B2-5, same-condition models again underperformed, whereas TQFMDCAE achieved the highest CI and CAFLAE delivered the second-best HI construction performance. Finally, a similar pattern was observed in B3-5: DCAE and SMSAE displayed relatively low performance, whereas CAFLAE achieved the best CI of 0.7331.

Models trained under the same-condition setting generally show lower HI construction performance and larger variability across operating conditions. In contrast, cross-condition models, such as TQFMDCAE and HCPTSCAE demonstrate clear advantages by aligning representations across domains. Notably, CAFLAE achieved high CI values in most tests, with an average improvement of 6.67% across existing methods. These findings indicate that the proposed framework generalizes effectively across diverse operating environments and mitigates domain shift during HI construction.

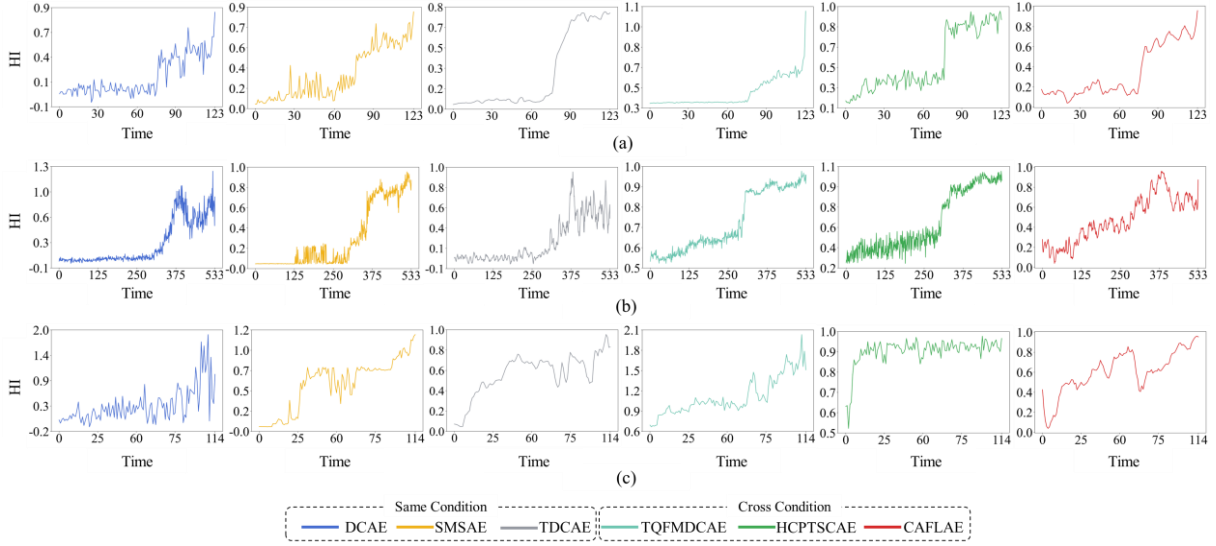

**Figure 9** Visualization of the constructed HIs on the XJTU-SY dataset for same condition and cross condition settings across (a) B1-1, (b) B2-3 and (c) B3-3. Each subplot compares the temporal HI trends produced by DCAE, SMSAE, TDCAE, TQFMDCAE, HCPTSCAE, and the CAFLAE

The constructed HI for B1-1, B2-3, and B3-5 under different operating conditions in the XJTU-SY dataset is shown in Figure 9, with the x- and y-axis representing the time and HI values, respectively. This enable the assessment of stability and degradation consistency over the run-to-failure process. Among same-condition models, DCAE generated unstable trajectories across all three bearings. In B1-1, pronounced irregular oscillations dominated the early stage, whereas a clear increasing trend only emerged later. In B2-3, the HI increases mainly near failure, but repeated fluctuations reduce interpretability. In B3-5, large-amplitude fluctuations and abrupt variations near failure compromise monitoring performance. SMSAE yields smoother HIs than DCAE and better reflects gradual degradation, although some fluctuations persist. TDCAE reduces early-stage noise and shows a monotonic increase in B1-1; however, it becomes unstable in the late stage for B2-3 and demonstrates increased variability in the mid-to-late stages for B3-5.

Cross-condition models generally yield clearer trajectories; however, their performance varies by bearing. TQFMDCAE demonstrates consistent, low-noise degradation for B1-1 and B2-3, whereas for B3-5 it yields higher HI amplitudes and ambiguous degradation near failure, increasing variance. Similarly, HCPTSCAE demonstrates low-noise trends for B1-1 and B2-3, yet in B3-5, the HI rises sharply from the outset, failing to capture gradual degradation. Conversely, CAFLAE consistently generates stable, monotonic degradation trajectories within the [0, 1] range for all bearings. For B1-1, it provides a clear, steadily increasing curve; for B2-3, a steeper mid-stage rise is observed, but the overall trajectory remains consistent until failure. For B3-5, the HI increases rapidly at the start but remains smooth and stable thereafter, avoiding the severe fluctuations observed in other methods. These results demonstrate that CAFLAE enhances HI construction compared with existing same- and cross-condition models, enabling superior performance.

**Table 8 Comparison of model complexities**

| Methods | Complexity | | |
|---|---|---|---|
| | Params # (M) | FLOPs # (M) | Inference time (ms) |
| DCAE | 1.0779 | 8.9924 | 1.4829 |
| SMSAE | 0.0584 | 47.9016 | 0.9836 |
| TDCAE | 1.2063 | 9.4862 | 2.9662 |
| TQFMDCAE | 0.4528 | 26.7853 | 2.1304 |
| HCPTSCAE | 0.4360 | 126.7097 | 2.0662 |
| CAFLAE | 0.4199 | 39.1120 | 1.8703 |

CAFLAE features the second fewest parameters (0.4199M) and the fourth lowest FLOPs (39.11M) among the models evaluated, offering notable efficiency during training on both the source and target datasets. Its inference time is 2.43 ms, 1.28× faster than the average inference time of 1.8703 ms, ensuring suitability for real-time prediction and condition monitoring. Moreover, because the KWS and XJTU-SY datasets are monitored at 360-s and 1-min intervals, respectively, the inference speed of CAFLAE pose no practical constraints for deployment in industrial deployment.

### 3.5.2 Validation of Learning Stability

This section presents three analyses to assess the training stability of the proposed framework. Specifically, we evaluated training stability through loss-curve analysis, and visualized the domain alignment status between the source and target domains using t-SNE.

Training stability is assessed by analyzing the evolution of training curves. PI-Control, a stability metric introduced by [43], quantifies the consistency of loss reduction during the final training phase and captures oscillations during optimization by computing the sample standard deviation of the loss values over the last $H$ epochs. First, the mean loss over the final $H$ epochs is defined as in Eq. (40),

$$\bar{\varepsilon} = \frac{1}{H} \sum_{e=E-H+1}^{E} \varepsilon_e \qquad (40)$$

where $E$ denotes the total number of training epochs, $L^e$ denotes the loss value at epoch $e$, and $H$ denotes the evaluation horizon used for stability assessment. PI-Control is then computed as the standard deviation of these losses, as expressed in Eq. (41).

$$PI - Control = \sqrt{\frac{1}{H-1} \sum_{e=E-H+1}^{E} (\varepsilon_e - \bar{\varepsilon})^2} \qquad (41)$$

Lower PI-Control values indicates reduced oscillations and greater stability, whereas higher values reflect increased fluctuations and reduced stability. In this study, PI-Control was computed using epochs after epoch 10.

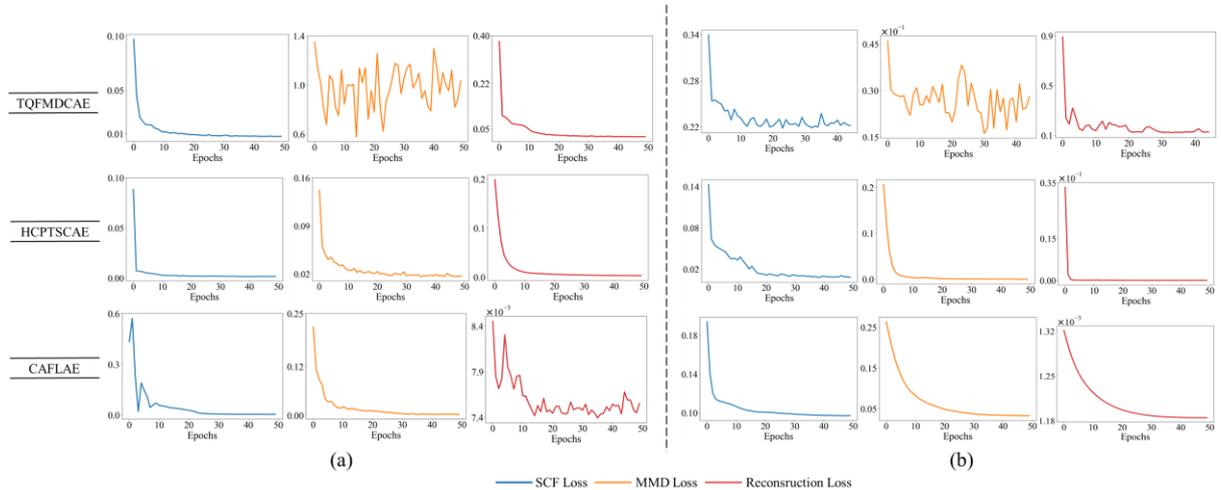

**Figure 10 (a) Training curve for RC1-2 and (b) for B1-1. These curves show the SCF, MMD, and reconstruction loss for each training epoch for TQFMDCAE, HCPTSCAE, and the proposed CAFLAE.**

**Table 9 PI-Control values of SCF, MMD, and reconstruction losses for the RC1-2 dataset. CAFLAE represents the most stable learning dynamics, with consistently lower PI-Control values compared with**

those of other models.

| Methods | SCF Loss | MMD Loss | Reconstruction Loss |
|---|---|---|---|
| TQFMDCAE | $2.331\times10^{-4}$ | $1350.979\times10^{-4}$ | $233.1\times10^{-6}$ |
| HCPTSCAE | **$0.606\times10^{-4}$** | $11.391\times10^{-4}$ | $94.2\times10^{-6}$ |
| CAFLAE | $1.125\times10^{-4}$ | **$0.361\times10^{-4}$** | **$68.1\times10^{-6}$** |

**Table 10 PI-Control values of SCF, MMD, and reconstruction losses for the B1-1 dataset. CAFLAE achieves the highest training stability, demonstrating the smallest fluctuations across all loss terms.**

| Methods | SCF Loss | MMD Loss | Reconstruction Loss |
|---|---|---|---|
| TQFMDCAE | $45.724\times10^{-4}$ | $41.411\times10^{-4}$ | $837.84\times10^{-6}$ |
| HCPTSCAE | $5.497\times10^{-4}$ | **$0.028\times10^{-4}$** | $5.51\times10^{-6}$ |
| CAFLAE | **$2.319\times10^{-4}$** | $5.331\times10^{-4}$ | **$0.1\times10^{-6}$** |

The training curves for the RC1-2 experiment are shown in Figure 10. The x- and y-axes represent the number of epochs and corresponding loss values, respectively. The overall trends of the SCF, MMD, and reconstruction losses for TQFMDCAE, HCPTSCAE, and CAFLAE during training are shown in the figure. While the cross-condition baselines demonstrate noticeable fluctuations, CAFLAE demonstrates smoother and more stable convergence across all loss components.

This trend is evident in PI-Control. On the KWS dataset, HCPTSCAE achieved the best PI-Control value for the SCF loss, whereas CAFLAE demonstrated superior stability for the MMD and reconstruction losses, indicating an advantage in overall training stability. On the XJTU-SY dataset, HCPTSCAE achieved the best PI-Control for the MMD loss, whereas CAFLAE ranked second. For the SCF and reconstruction losses, CAFLAE achieved the best results. Overall, CAFLAE consistently delivers competitive PI-Control values across the SCF, MMD, and reconstruction terms on both datasets, demonstrating stable optimization.

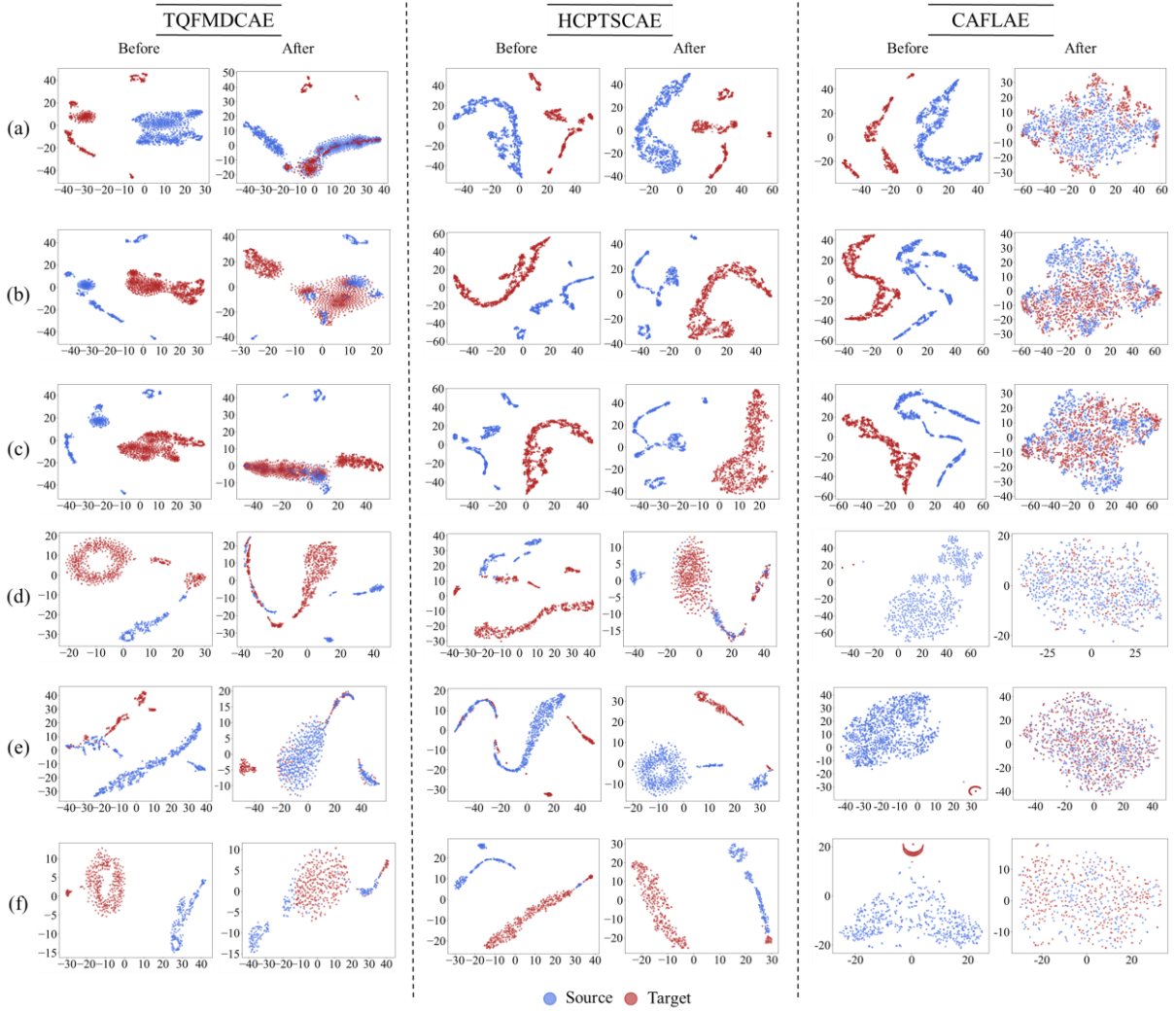

**Figure 11** t-SNE visualization of source and target domain features before and after training for the KWS dataset under the DA-based HI construction framework. Subfigures correspond to the following test conditions: (a) RC1-2, (b) RC1-3, (c) AC2-2, (d) B1-1, (e) B2-3, and (f) B3-5.

    The t-SNE visualizations of the source and target domains before and after training for all experimental settings are shown in Figure 11 (a)–(f). Initially, the domains were clearly separated owing to differing degradation patterns. After training, TQFMDCAE achieves partial alignment, with some regions of the source and target distributions moving closer—such as the target distribution in RC1-2 partially following the source. However, a substantial portion of source and target samples remains separated even after training, indicating unresolved domain shift. HCPTSCAE generally demonstrates weaker alignment performance, except for B1-1; most conditions display fragmented and separated distributions post-training, consistent with previously observed MMD loss instability and suggesting inadequate learning of degradation representations. In contrast, CAFLAE consistently achieves dense overlap and alignment between source and target distributions across all conditions after training. The proposed framework reliably aligns degradation representations across domains. The improvement stems from synchronizing degradation stages during sampling and preventing domain loss computation on samples with heterogeneous degradation patterns. Consequently, the combination of CAFLAE and DSSBS achieves high-quality and consistent domain alignment across all settings (a)–(f), demonstrating effective DA-based HI construction.

### 3.5.3 ERF-Based Large Kernel Validation

Industrial time-series signals often span long temporal ranges owing to high sampling frequencies and long monitoring periods, introducing challenges, such as long-range dependencies, noise, and non-stationarity [6]. Capturing long-range dependencies is therefore critical. This section evaluates whether the large kernels in CAFLAE effectively extract such long-range dependencies using ERF analysis [28,30].

Following [44], ERF is quantified by computing gradients of a scalar output with respect to the input, measuring the sensitivity of each input time point. For a convolutional input $X \in \mathbb{R}^{C \times N \times P}$ an feature tensor $Y \in \mathbb{R}^{P \times N}$ with temporal length $N$, we first compute the time-wise mean to obtain a scalar output $q = \text{mean}(Y)$.

The input gradient $\nabla_X q$ is then computed. Finally, the sensitivity vector for each time step is computed as the mean of the absolute gradients over the channel dimension, as defined in Eq. (42):

$$ERF(t) = \frac{1}{CP} \sum_{c=1}^{C} \sum_{p=1}^{P} \left| \frac{\partial q}{\partial X_{c,t,p}} \right| \quad (42)$$

where $X_{(c,t)}$ denotes the value at time $t$ in channel $c$, $p$ and $q$ denotes the scalar output. This computation employed the convolutional outputs. ERF was computed on five randomly selected mini-batches and averaged.

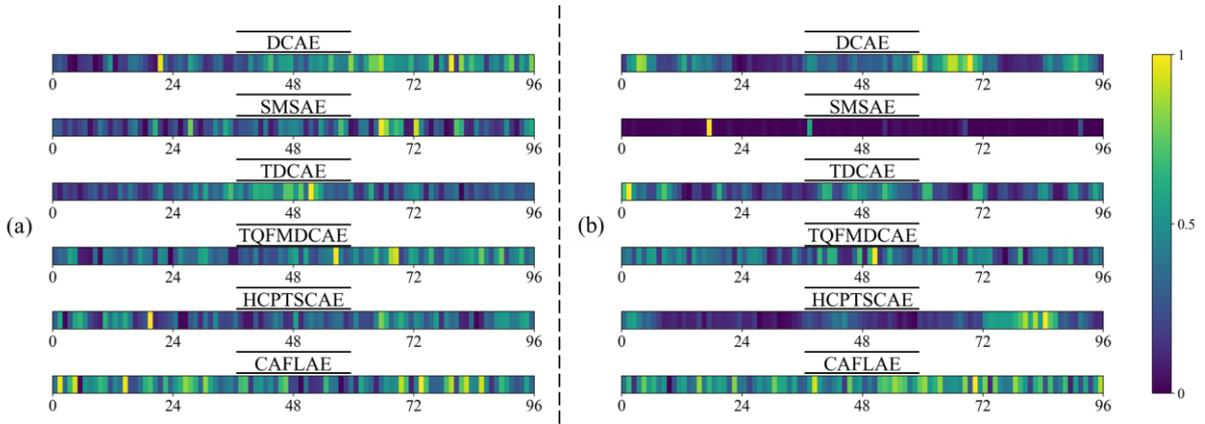

**Figure 12 Visualization of the ERF on the (a) RC1-2 and (b) B1-1 dataset.**

The ERF of the compared models on RC1-2 and B1-1 is shown in Figure 12. The x-axis denotes the batch dimension, whereas the y-axis denotes the time axis, where brighter pixels indicate higher sensitivity of the model to the corresponding time points. Distinct differences were observed among models in terms of the broad and consistent utilization of temporal information. DCAE, SMSAE, and TDCAE demonstrate highly localized and fragmented ERFs, concentrating sensitivity on narrow temporal regions. This indicates an over-reliance on limited segments of the input segments and an inability to capture the overall degradation pattern. SMSAE, in particular, focuses its ERF on a few temporal regions owing to its self-attention mechanism, making it less suitable for long time-series feature extraction. TQFMDCAE and HCPTSCAE, which incorporate multi-scale structures, achieve broader ERFs; however, their ERF distributions remain uneven and lack temporal continuity along. Notably, for B1-1, the ERF of HCPTSCAE is almost exclusively activated around the 72nd time step, with noticeable discontinuities elsewhere, highlighting structural limitations in integrating long-term degradation information. Conversely, CAFLAE consistently produces broad, smooth, and uniformly distributed ERFs across both RC1-2 and B1-1, maintaining stable sensitivity throughout the temporal range without overemphasizing specific segments. This is attributed to the large-kernel PMTC block, which expands the ERF and addresses the localized patterns typical of conventional CNN-based architectures [28]. Overall, the ERF analysis demonstrates that CAFLAE most effectively leverages long-term degradation patterns, responds globally over time, and learns clearer degradation representations.

## 4. Parameter Sensitivity Analysis

This section presents a parameter sensitivity analysis for each module of the proposed DSSBS and CAFLAE framework. Specifically, Section 4.1 examines the sensitivity of the CPD algorithm used in DSSBS while considering kernel-size variations in CAFLAE. Section 4.2 evaluates the effectiveness of dynamic weight averaging (DWA).

### 4.1 Sensitivity to CPD Algorithms

To evaluate the effectiveness of the CPD algorithm within DSSBS, we conducted experiments using binary segmentation [39], bottom-up segmentation [40], dynamic programming [41], and KCP [45]. The compared variants are defined as follows:

- CAFLAE-BS: DSSBS implemented using binary segmentation
- CAFLAE-BU: DSSBS implemented using bottom-up segmentation
- CAFLAE-DP: DSSBS implemented using dynamic programming
- CAFLAE: DSSBS implemented using KCP

**Table 11 CPD algorithm sensitivity on the KWS dataset**

| Methods | RC1-2 | RC1-3 | AC2-2 |
|---|---|---|---|
| CAFLAE-BS | 0.7141 | 0.7265 | 0.6933 |
| CAFLAE-BU | 0.7068 | 0.7261 | 0.6286 |
| CAFLAE-DP | 0.6992 | 0.7271 | 0.6691 |
| CAFLAE | **0.7406** | **0.7416** | **0.7291** |

**Table 12 CPD algorithm sensitivity on the XJTU-SY dataset**

| Methods | B1-1 | B1-2 | B2-3 | B2-4 | B2-5 | B3-5 |
|---|---|---|---|---|---|---|
| CAFLAE-BS | 0.6832 | 0.7445 | 0.7493 | 0.8427 | 0.6821 | 0.6634 |
| CAFLAE-BU | 0.6973 | 0.7267 | **0.7616** | 0.8374 | 0.6958 | 0.6898 |
| CAFLAE-DP | 0.7180 | 0.7169 | 0.7357 | **0.8447** | 0.7005 | 0.6887 |
| CAFLAE | **0.7706** | **0.7794** | 0.7263 | 0.8176 | **0.7312** | **0.7331** |

The results reported in Tables 11 and 12 demonstrate the sensitivity of DSSBS to different CPD algorithms. For the KWS dataset, CAFLAE with KCP consistently outperformed other variants. Because the KWS data comprises two input channels, kernel-based KCP operating in a kernel-induced feature space more expressively captures multi-channel degradation patterns, making it particularly suitable when degradation occurs across multiple correlated channels. For the XJTU-SY dataset, sensitivity to the CPD algorithm varied across test bearings. CAFLAE-BU performed best for B2-3, whereas CAFLAE-DP performed best for B2-4. However, DSSBS with KCP achieved the highest performance for most bearings, including B1-1, B1-2, B2-5, and B3-5. Overall, kernel-based CPD offers the most reliable performance across diverse operating conditions. Overall, kernel-based CPD emerged as the most robust and stable CPD approach within DSSBS, consistently delivering strong performance across diverse datasets and degradation patterns. Therefore, employing KCP as the CPD algorithm in the proposed methodology is well justified.

### 4.2 Effectiveness of DWA

The loss function of the proposed framework is formulated as a weighted sum of multiple objective terms, where the weights are dynamically adjusted using DWA. Conventional methods typically assign fixed weights to each objective term. Such fixed-weight strategies have the following limitations:

- The number of weight combinations increases with the number of tasks.
- Extensive search is required to identify suitable weights.
- Optimal weights are often dataset-dependent, limiting the generalizability of fixed-weight schemes.

To examine the practical impacts of these weighting strategies, we conducted comparative experiments using both the proposed DWA-based dynamic weighting and conventional fixed-weight schemes. DWA adjusts task weights during training based on the magnitude and relative change of each task loss, promoting more stable optimization and reducing the need for manual hyperparameter tuning. Accordingly, this section compares (1) fixed-weight groups (G1–G5) and (2) the proposed DWA-based weighting strategy, evauating their impact on HI construction performance. Specifically we assessed the five manually specified loss-weight groups (G1–G5) referenced in [16], as well as the dynamically adjusted DWA method.

**Table 13 Summary of the parameter sensitivity results comparing fixed loss-weight combinations and the dynamically adjusted DWA for HI construction on the KWS and XJTU-SY datasets.**

| Test | Group | Weights | | | Metric | | | |
| --- | --- | --- | --- | --- | --- | --- | --- | --- |
| | | $\lambda_1$ | $\lambda_2$ | $\lambda_3$ | Mon | Cor | Rob | CI |
| RC1-2 | G1 | 1 | 0.5 | 0.5 | 0.4908 | <u>0.8229</u> | <u>0.8765</u> | 0.7061 |
| | G2 | 1 | 0.6 | 0.6 | <u>0.4988</u> | 0.8144 | 0.8666 | 0.7038 |
| | G3 | 1 | 0.7 | 0.7 | 0.4978 | 0.8197 | 0.8718 | <u>0.7066</u> |
| | G4 | 1 | 0.8 | 0.8 | 0.4961 | 0.8165 | 0.8676 | 0.7037 |
| | G5 | 1 | 0.9 | 0.9 | 0.4925 | 0.8212 | 0.8671 | 0.7035 |
| | DWA | | | | **0.5055** | **0.8719** | **0.9226** | **0.7406** |
| B1-1 | G1 | 1 | 0.5 | 0.5 | 0.5418 | 0.8234 | 0.9300 | 0.7427 |
| | G2 | 1 | 0.6 | 0.6 | 0.5405 | 0.8030 | **0.9467** | 0.7411 |
| | G3 | 1 | 0.7 | 0.7 | <u>0.5661</u> | 0.7989 | 0.9238 | 0.7432 |
| | G4 | 1 | 0.8 | 0.8 | 0.5451 | 0.8299 | 0.9313 | 0.7464 |
| | G5 | 1 | 0.9 | 0.9 | 0.5477 | <u>0.8489</u> | 0.9282 | <u>0.7522</u> |
| | DWA | | | | **0.5702** | **0.8634** | <u>0.9451</u> | **0.7706** |

A summary of the impact of DWA on HI construction using the RC1-2 and B1-1 datasets is provided in Table 23. For RC1-2, some fixed-weight groups (G1–G5) demonstrated partial improvements on specific metrics. However, DWA consistently outperformed all groups across Mon, Cor, Rob, and CI. Notably, the CI score increased from 0.7061 (G1) to 0.7706 (DWA), a 9.2% increase, demonstrating the effectiveness of DWA in balancing multiple loss terms and enhancing training stability. Similar resilts were observed for B1-1. Although G2 achieved a relatively high Rob value, DWA achieved the highest CI when considering all metrics. Importantly, DWA eliminates the extensive manual tuning required to optimize multi-objective loss functions under fixed-weight schemes. By dynamically updating weights based on the relative change rates among losses, DWA prevents overemphasis on any single loss term and promotes a more balanced optimization trajectory. Overall, DWA enhances both stability and performance of HI construction, offering significant advantages in DA scenarios requiring joint optimization of multiple losses.

## 5. Ablation Study

This section comprehensively summarizes ablation experiments evaluating the effectiveness of the proposed method. Section 5.1 compares model performance with and without the DSSBS mechanism. Section 5.2 validates the use of large kernels in PMTC by analyzing different multi-scale kernel sizes. Section 5.3 evaluates the impact of the multi-scale design in CAFLAE by comparing single-kernel and multi-scale configurations. Section 5.4 analyzes how the proposed cross-attention-based domain fusion affects alignment between the source and target domains.

### 5.1 Analysis of the Effectiveness of DSSBS

To assess the effectiveness of DSSBS, we conducted an ablation analysis by comparing results with and without DSSBS.

**Table 14 Experimental results on the KWS dataset DSSBS ablation study**

| Methods | RC1-2 | RC1-3 | AC2-2 |
| --- | --- | --- | --- |
| w/o DSSBS | 0.6367 | 0.5949 | 0.5489 |
| DSSBS | **0.7406** | **0.7416** | **0.7291** |

**Table 15 Experimental results on XJTU-SY dataset DSSBS ablation study**

| Methods | B1-1 | B1-2 | B2-3 | B2-4 | B2-5 | B3-5 |
|---|---|---|---|---|---|---|
| w/o DSSBS | 0.6222 | 0.7095 | 0.6851 | 0.6881 | 0.6629 | 0.5735 |
| DSSBS | **0.7706** | **0.7794** | **0.7263** | **0.8176** | **0.7312** | **0.7331** |

The ablation results on the KWS and XJTU-SY datasets are listed in Tables 14 and 15, respectively. Removing DSSBS significantly degraded overall HI construction performance in terms of CI, reducing the average CI by 14.4% on KWS and 17.4% on XJTU-SY. This substantial reduction indicates that proper domain alignment cannot be achieved when domain losses are computed from randomly sampled source and target mini-batches. This is primarily because degradation stages are not synchronized across different equipment, leading to alignment under stage-mismatched conditions. Therefore, DSSBS, which synchronizes source and target samples by degradation stage, is therefore essential for achieving stable domain adaptation.

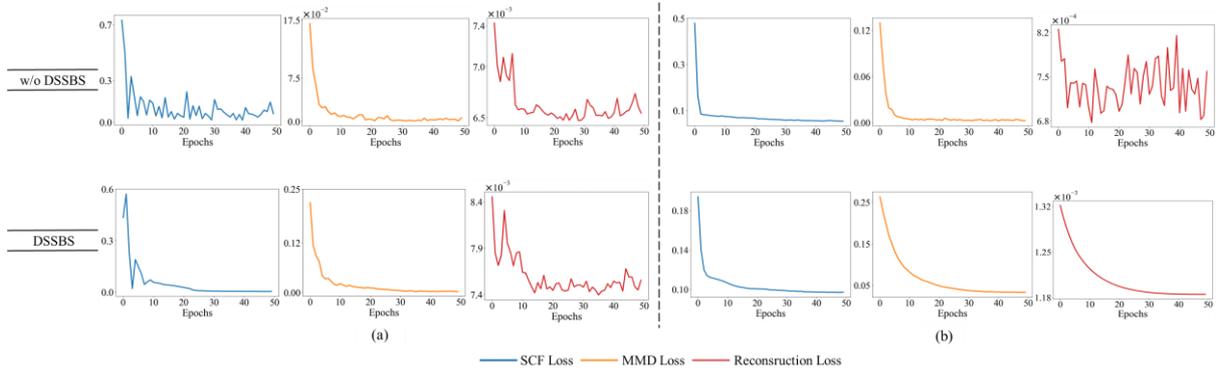

**Figure 13 Learning curves with and without DSSBS on (a) RC1-2 and (b) B1-1.**

**Table 16 PI-Control values of SCF, MMD, and reconstruction losses for the RC1-2 dataset.**

| Methods | SCF Loss | MMD Loss | Reconstruction Loss |
|---|---|---|---|
| w/o DSSBS | $300.372 \times 10^{-4}$ | $14.298 \times 10^{-4}$ | $68.8 \times 10^{-6}$ |
| DSSBS | $\mathbf{1.125 \times 10^{-4}}$ | $\mathbf{0.361 \times 10^{-4}}$ | $\mathbf{68.1 \times 10^{-6}}$ |

**Table 17 PI-Control values of SCF, MMD, and reconstruction losses for the B1-1 dataset.**

| Methods | SCF Loss | MMD Loss | Reconstruction Loss |
|---|---|---|---|
| w/o DSSBS | $7.735 \times 10^{-4}$ | $6.316 \times 10^{-4}$ | $3 \times 10^{-6}$ |
| DSSBS | $\mathbf{2.319 \times 10^{-4}}$ | $\mathbf{5.331 \times 10^{-4}}$ | $\mathbf{0.1 \times 10^{-6}}$ |

The training curves for RC1-2 and B1-1 are shown in Figure 13, where the x-axis and y-axis denote the number of epochs and corresponding loss values, respectively. A summary of PI-Control values computed over the final 10-epoch window is presented in Tables 16 and 17. These results reinforce the aforementioned observations. For both RC1-2 and B1-1, models trained without DSSBS demonstrated higher PI-Control values, indicating unstable training characterized by large fluctuations in the loss curves. This instability results from inconsistent sampling of degradation stages across domains, yielding incorrect domain-alignment losses. In contrast, models trained with DSSBS achieved lower PI-Control values across all loss components—including SCF, MMD, and reconstruction losses—and their training curves converged more smoothly and rapidly. These findings demonstrate that DSSBS enables stable domain alignment and robust HI construction.

## 5.2 Analysis of the Multi-scale Kernel Size

Small kernels in 1D-CNN are typically used for convolutions on frequency-domain or time-series data. To evaluate the impact of large kernels in CAFLAE, we developed PMTC variants with different kernel configurations and performed comparative analyses:

- CAFLAE-S: PMTC using a small-kernel set $\mathbb{K} = \{3, 5, 7\}$

- CAFLAE-M: PMTC using a medium-kernel set $\mathbb{K} = \{11, 15, 21\}$
- CAFLAE: PMTC using a large-kernel set $\mathbb{K} = \{13, 23, 31\}$

**Table 18 Ablation results of kernel size on the KWS dataset**

| Methods | RC1-2 | RC1-3 | AC2-2 |
| --- | --- | --- | --- |
| CAFLAE-S | 0.7003 | 0.7300 | 0.6696 |
| CAFLAE-M | 0.6994 | 0.7312 | 0.6742 |
| CAFLAE | **0.7406** | **0.7416** | **0.7291** |

**Table 19 Ablation results of kernel size on the XJTU-SY dataset**

| Methods | B1-1 | B1-2 | B2-3 | B2-4 | B2-5 | B3-5 |
| --- | --- | --- | --- | --- | --- | --- |
| CAFLAE-S | 0.7412 | 0.7246 | 0.7266 | 0.8006 | 0.6310 | 0.6868 |
| CAFLAE-M | 0.7339 | 0.7078 | **0.7375** | 0.8005 | 0.6128 | 0.6993 |
| CAFLAE | **0.7706** | **0.7794** | 0.7263 | **0.8176** | **0.7312** | **0.7331** |

The ablation results for kernel sizes in the multi-scale convolution structure are listed in Tables 18 and 19. On the KWS dataset, the large-kernel configuration consistently achieved the best CI performance across all tests. For RC1-2, RC1-3, and AC2-2, the full large-kernel setting achieved CI scores of 0.7406, 0.7416, and 0.7291, respectively, surpassing both small-kernel (CAFLAE-S) and medium-kernel (CAFLAE-M) variants. A similar trend was observed on the XJTU-SY dataset. For example, on B1-1, B1-2, and B2-4, CAFLAE achieved CI scores of 0.7706, 0.7794, and 0.8176, respectively, outperforming CAFLAE-S and CAFLAE-M by 0.028 and 0.045 on average. Notably, the large-kernel configuration also yielded the best results in the B2-3 experiment, indicating robustness to variations in operating conditions. Overall, these findings demonstrate that a large-kernel-based multi-scale architecture enhances HI construction performance.

### 5.3 Multi-Scale Effectiveness Analysis of PMTC

This section presents an ablation study to evaluate the effectiveness of the multi-scale design proposed in CAFLAE. We compared the full multi-scale PMTC against a single-kernel PMTC configuration:

- CAFLAE-SK: PMTC with $\mathbb{K} = \{13\}$
- CAFLAE: PMTC with $\mathbb{K} = \{13, 23, 31\}$

**Table 20 Ablation results of the multiscale design on the KWS dataset.**

| Methods | RC1-2 | RC1-3 | AC1-2 |
| --- | --- | --- | --- |
| CAFLAE-SK | 0.7113 | 0.7069 | 0.5720 |
| CAFLAE | **0.7406** | **0.7416** | **0.7291** |

**Table 21 Ablation results of the multiscale design on the XJTU-SY dataset.**

| Methods | B1-1 | B1-2 | B2-3 | B2-4 | B2-5 | B3-5 |
| --- | --- | --- | --- | --- | --- | --- |
| CAFLAE-SK | 0.7092 | 0.6639 | 0.6978 | **0.8258** | 0.5305 | 0.5833 |
| CAFLAE | **0.7706** | **0.7794** | **0.7263** | 0.8176 | **0.7312** | **0.7331** |

The multi-scale ablation results on the KWS and XJTU-SY datasets are summarized in Tables 20 and 21, respectively. Removing the multi-scale module (CAFLAE-SK) significantly degraded performance, decreasing CI by 11.1% and 13.66%, respectively. This reduction occurs because a single-kernel configuration extracts degradation patterns only at one temporal resolution, resulting in limited feature representation. As equipment degradation signals demonstrate long-range temporal dependencies, integrating kernels at multiple scales is necessary for comprehensively capturing diverse temporal patterns. Therefore, the multi-scale structure is crucial for learning complex degradation patterns and significantly enhances HI construction performance in CAFLAE.

### 5.4 Effectiveness Analysis of Cross Attention

Finally, we evaluated the effectiveness of the cross-attention module in the proposed CAFLAE by conducting an ablation study that compares a model without cross attention to the full model with cross attention under identical training conditions.

**Table 22 Ablation results on the impact of cross attention for the KWS dataset**

| Methods | RC1-2 | RC1-3 | AC2-2 |
| --- | --- | --- | --- |
| w/o Cross Attention | 0.6788 | **0.7425** | 0.7101 |
| w Cross Attention | **0.7406** | 0.7416 | **0.7291** |

**Table 23 Ablation results on the impact of cross attention for the XJTU-SY dataset**

| Methods | B1-1 | B1-2 | B2-3 | B2-4 | B2-5 | B3-5 |
| --- | --- | --- | --- | --- | --- | --- |
| w/o Cross Attention | 0.7220 | 0.7308 | 0.7217 | 0.6251 | 0.6158 | 0.6694 |
| w Cross Attention | **0.7706** | **0.7794** | **0.7263** | **0.8176** | **0.7312** | **0.7331** |

The ablation results on the KWS and XJTU-SY datasets are listed in Tables 22 and 23, respectively. On the KWS dataset, the model with cross attention consistently achieved higher CI scores in most experiments, except for RC3. Notably, CI increased by 9.1% on RC1-2 and by 2.68% on AC2-2, demonstrating that cross-attention facilitates richer cross-domain information exchange and more robust degradation modeling. A similar trend was observed on the XJTU-SY dataset, where the model with cross-attention achieved performance improvements across all bearing conditions. CI improved by 30.8% on B2-4, indicating that cross-attention effectively fuses degradation patterns across domains, enabling higher-performance HI construction.

## 6. Conclusion

This study proposed a high-performance domain-adaptive HI construction framework that addressed inaccurate domain-loss computation from mixed degradation stages using DSSBS, and enhances long-horizon temporal feature extraction through CAFLAE with a large-kernel neural architecture. DSSBS synchronized degradation stages between source and target domains during training, preventing incorrect loss computation that stemmed from the random mixing of heterogeneous degradation stages. This sampling strategy stabilized domain alignment and enhanced robustness throughout the training process. CAFLAE integrated a PMTC structure with an expanded receptive field and a cross-attention mechanism, enabling effective capture of long-range temporal dependencies and complementary cross-domain information. These components mitigated the core limitations of existing DA-based HI methods, supporting more reliable HI construction in real industrial environments.

Validation on the KWS and XJTU-SY bearing dataset demonstrated that DSSBS and CAFLAE consistently outperform existing methods in HI construction performance. Training loss curves and t-SNE visualizations before and after training confirmed significant improvements in both training stability and domain alignment over baseline models. Furthermore, ERF analysis indicated that CAFLAE secured a broad ERF and effectively captured long-term degradation patterns over the entire operating period through the large-kernel structure. Furthermore, ablation studies validated that each component of DSSBS and CAFLAE is essential for enhancing DA stability, improving long-horizon temporal feature extraction, and boosting HI construction performance.

However, further research is required to improve generalization across diverse equipment types and operating conditions. In this study, degradation stages were identified using RMS; however, integrating both time- and frequency-domain features could provide richer degradation descriptors and enable more refined stage identification. Moreover, although PMTC was adopted to enhance long-horizon temporal feature extraction, exclusive reliance on large kernels may overlook short-term patterns. Future work should investigate architectures that combine large and small kernels to jointly capture both short- and long-term degradation characteristics.

**Acknowledgments**

This research was supported by the Korea Research Institute for Defense Technology Planning and Advancement (KRIT) Grant funded by the Defense Acquisition Program Administration (DAPA) (KRIT-CT-22-081, Weapon System CBM+ Research Center, 50) and the Institute of Information & Communications Technology Planning &

Evaluation(IITP)-Innovative Human Resource Development for Local Intellectualization program grant funded by the Korea government(MSIT)(IITP-2025-RS-2020-II201791, 50)

**Data availability**

- The XJTU-SY dataset is available from the authors upon reasonable request.
- The KWS weapon system dataset is not publicly available owing to security restrictions.

**Appendix**

**Table 24 Dataset configurations for XJTU-SY under Cross Condition-b**

| Type | Test | Dataset | | | Models |
|---|---|---|---|---|---|
| | | Train | | Condition | |
| | | Source (Labeled) | Target (Unlabeled) | | |
| Cross Condition-b | B1-1, B1-2 | B3-3 | B1-1, B1-2 | 3 → 1 | TQFMDCAE |
| | B2-3, B2-4, B2-5 | B3-3 | B2-1, B2-2 | 3 → 2 | HCPTSCAE |
| | B3-5 | B2-1, B2-2 | B3-3 | 2 → 3 | CAFLAE |

**Table 25 Comparison of HI construction performance on the XJTU-SY Dataset-b**

| Test | Method | Metric | | | |
|---|---|---|---|---|---|
| | | Mon | Cor | Rob | CI |
| Bearing1-1 | DCAE | 0.5494 | 0.8343 | 0.5003 | 0.6202 |
| | SMSAE | 0.5524 | **0.8617** | 0.5659 | 0.6523 |
| | TDCAE | 0.6808 | 0.8571 | 0.5184 | 0.6880 |
| Bearing1-1-b | TQFMDCAE | 0.5565 | 0.6442 | 0.9063 | 0.6878 |
| | HCPTSCAE | 0.4775 | 0.6451 | 0.8691 | 0.6452 |
| | CAFLAE | **0.5785** | 0.8565 | **0.9169** | **0.7634** |
| Bearing1-2 | DCAE | 0.5390 | 0.8255 | 0.6400 | 0.6553 |
| | SMSAE | 0.5526 | 0.8742 | 0.6790 | 0.6870 |
| | TDCAE | **0.6484** | **0.9024** | 0.6683 | 0.7306 |
| Bearing1-2-b | TQFMDCAE | 0.5519 | 0.7753 | 0.8837 | 0.7185 |
| | HCPTSCAE | 0.4738 | 0.5306 | **0.8899** | 0.6157 |
| | CAFLAE | 0.5845 | 0.8142 | 0.8503 | **0.7331** |
| Bearing2-3 | DCAE | 0.5096 | 0.8263 | 0.4167 | 0.5767 |
| | SMSAE | **0.5210** | **0.8600** | 0.4679 | 0.6068 |
| | TDCAE | 0.5165 | 0.8480 | 0.4432 | 0.5939 |
| Bearing2-3-b | TQFMDCAE | 0.4817 | 0.7586 | 0.8529 | 0.6761 |
| | HCPTSCAE | 0.4518 | 0.6695 | 0.8201 | 0.6276 |
| | CAFLAE | 0.5125 | 0.7427 | **0.9373** | **0.7090** |
| Bearing2-4 | DCAE | 0.6204 | 0.7738 | 0.3276 | 0.5786 |
| | SMSAE | 0.6041 | **0.8029** | 0.4208 | 0.6087 |
| | TDCAE | **0.6561** | 0.7554 | 0.3692 | 0.5998 |
| Bearing2-4-b | TQFMDCAE | 0.5672 | 0.7180 | 0.8435 | **0.6953** |
|  | HCPTSCAE | 0.4359 | 0.6974 | 0.8836 | 0.6487 |
|  | CAFLAE | 0.5177 | 0.3609 | **0.9739** | 0.6075 |
| Bearing2-5 | DCAE | 0.5245 | 0.8231 | 0.5492 | 0.6215 |
|  | SMSAE | 0.5435 | **0.9090** | 0.6869 | 0.6962 |
|  | TDCAE | 0.5269 | 0.8364 | 0.6585 | 0.6592 |
| Bearing2-5-b | TQFMDCAE | **0.5472** | 0.8233 | 0.8906 | **0.7330** |
|  | HCPTSCAE | 0.4880 | 0.5596 | 0.8600 | 0.6211 |
|  | CAFLAE | 0.4866 | 0.7091 | **0.9021** | 0.6780 |
| Bearing3-5 | DCAE | 0.5143 | 0.4235 | 0.5253 | 0.4903 |
|  | SMSAE | 0.5679 | **0.7882** | 0.7714 | 0.6950 |
|  | TDCAE | **0.6301** | 0.6935 | 0.7717 | 0.6916 |
| Bearing3-5-b | TQFMDCAE | 0.5593 | 0.7662 | 0.8101 | **0.6966** |
|  | HCPTSCAE | 0.5183 | 0.5868 | 0.9059 | 0.6551 |
|  | CAFLAE | 0.5326 | 0.6063 | **0.9834** | 0.6900 |

**Table 26 Final Selected Hyperparameter configurations for each Model**

| Model | Methods | Selected Hyperparameters | |
|---|---|---|---|
|  |  | Model architecture | Model Training |
| KWS | DCAE | $N_e = 6; N_d = 6;$ | $\alpha = 1; \lambda_1 = 0.5; \lambda_2 = 0.5$ |
|  | SMSAE | $N_e = 6; N_d = 6; d = 64,$ | $\alpha = 1; \lambda_1 = 0.5; \lambda_2 = 0.5$ |
|  | TDCAE | $N_e = 6; N_d = 6; d = 128$ | $\alpha = 1; \lambda_1 = 0.5; \lambda_2 = 0.5$ |
|  | TQFMDCAE | $N_e = 3; N_d = 4; k = \{3, 5, 7\}$ | $\alpha = 1; \lambda_1 = 0.5; \lambda_2 = 0.5; \lambda_3 = 0.5$ |
|  | HCPTSCAE | $N_e = 3; N_d = 3; d = 64;$ $k = \{3, 5, 7\}$ | $\alpha = 1; \lambda_1 = 0.5; \lambda_2 = 0.5; \lambda_3 = 0.5;$ $\lambda_4 = 0.5; \lambda_5 = 0.5$ |
|  | CAFLAE | $N_e = 3; N_d = 3; d = \{32, 64\},$ $C = \{32, 64\}; k = \{13, 23, 31\},$ $r = \{4\}$ | $\alpha = 1; \lambda_1 = 0.5; \lambda_2 = 0.5; \lambda_3 = 0.5$ $penalty = 10; kernel = RBF$ |
| XJTU-SY | DCAE | $N_e = 6; N_d = 6;$ | $\alpha = 1; \lambda_1 = 0.5; \lambda_2 = 0.5$ |
|  | SMSAE | $N_e = 6; N_d = 6; d = 64,$ | $\alpha = 1; \lambda_1 = 0.5; \lambda_2 = 0.5$ |
|  | TDCAE | $N_e = 6; N_d = 6; d = 128$ | $\alpha = 1; \lambda_1 = 0.5; \lambda_2 = 0.5$ |
|  | TQFMDCAE | $N_e = 3; N_d = 4; k = \{3, 5, 7\}$ | $\alpha = 1; \lambda_1 = 0.5; \lambda_2 = 0.5; \lambda_3 = 0.5$ |
|  | HCPTSCAE | $N_e = 3; N_d = 3; d = 64;$ $k = \{3, 5, 7\}$ | $\alpha = 1; \lambda_1 = 0.5; \lambda_2 = 0.5; \lambda_3 = 0.5;$ $\lambda_4 = 0.5; \lambda_5 = 0.5$ |
|  | CAFLAE | $N_e = 3; N_d = 3; d = \{32, 64\},$ $C = \{32, 64\}; k = \{13, 23, 31\},$ $r = \{4\}$ | $\alpha = 1; \lambda_1 = 0.5; \lambda_2 = 0.5; \lambda_3 = 0.5$ $penalty = 10; kernel = RBF$ |